\documentclass[10pt, a4paper]{article}

\usepackage{lrec-coling2024} 

\usepackage{booktabs}
\usepackage{multirow}
\usepackage{graphicx}
\usepackage{caption}
\usepackage{subcaption}
\usepackage{amsfonts}
\usepackage{algorithm}
\usepackage{algorithmic}
\usepackage{amssymb}
\usepackage{pifont}
\usepackage{xcolor}

\definecolor{deepblue}{rgb}{0,0,0.5}
\definecolor{officeblue}{RGB}{0,102,204}
\definecolor{deepred}{rgb}{0.6,0,0}
\definecolor{deepgreen}{rgb}{0,0.5,0}
\definecolor{mybrickred}{RGB}{182,50,28}
\definecolor{fillcolor}{RGB}{216,217,252}

\newcommand{\sptk}[1]{\texttt{[#1]}}
\renewcommand{\algorithmiccomment}[1]{\bgroup\hfill $\triangleright$ ~#1\egroup}

\newcommand{\xmark}{{\color{deepred}\ding{55}}}

\definecolor{color_m}{RGB}{72,117,170}
\definecolor{color_f}{RGB}{201,89,72}
\definecolor{color_c}{RGB}{230,230,230}
\definecolor{color_e}{RGB}{100,155,74}
\definecolor{ccon}{HTML}{fee9d4}
\definecolor{cood}{HTML}{d8f0d3}
\definecolor{cid}{HTML}{dae8f5}
\definecolor{gg}{HTML}{e2f0cb}

\usepackage{amsmath,amsfonts,bm}

\def\eqref#1{equation~\ref{#1}}

\def\1{\bm{1}}

\def\vg{{\bm{g}}}

\def\vx{{\bm{x}}}
\def\vy{{\bm{y}}}

\def\mA{{\bm{A}}}

\def\mG{{\bm{G}}}

\def\mI{{\bm{I}}}

\def\mM{{\bm{M}}}

\def\mS{{\bm{S}}}

\def\mW{{\bm{W}}}
\def\mX{{\bm{X}}}

\DeclareMathAlphabet{\mathsfit}{\encodingdefault}{\sfdefault}{m}{sl}
\SetMathAlphabet{\mathsfit}{bold}{\encodingdefault}{\sfdefault}{bx}{n}

\newcommand{\R}{\mathbb{R}}

\newcommand{\our}{\textsc{SmartTrim}}

\title{\our{}: Adaptive Tokens and Attention Pruning for Efficient Vision-Language Models}

\name{
  \begin{tabular}{c}
    Zekun Wang$^{1*}$, Jingchang Chen$^{1*}$, Wangchunshu Zhou$^{2}$, Haichao Zhu\\
    Jiafeng Liang$^{1}$, Liping Shan$^{3}$, Ming Liu$^{1, 4}$, Dongliang Xu$^3$, Qing Yang$^3$, \\
    Bing Qin$^{1,4}$
  \end{tabular}
}

\address{$^{1}$Harbin Institute of Technology ~~$^{2}$ ETH Zurich\\
$^{3}$Du Xiaoman (Beijing) Science Technology Co., Ltd ~~$^{4}$ Peng Cheng Laboratory \\
\{zkwang, jcchen, mliu, qinb\}@ir.hit.edu.cn\\}

\abstract{
Despite achieving remarkable performance on various vision-language tasks, Transformer-based Vision-Language Models (VLMs) suffer from redundancy in inputs and parameters, significantly hampering their efficiency in real-world applications.
Moreover, the degree of redundancy in token representations and model parameters, such as attention heads, varies significantly for different inputs.
In light of the challenges, we propose~\our{}, an adaptive acceleration framework for VLMs, which adjusts the computational overhead per instance.
Specifically, we integrate lightweight modules into the original backbone to identify and prune redundant token representations and attention heads within each layer.
Furthermore, we devise a self-distillation strategy to enhance the consistency between the predictions of the pruned model and its fully-capacity counterpart.
Experimental results across various vision-language tasks consistently demonstrate that \our{} accelerates the original model by $2$-$3$ times with minimal performance degradation, highlighting the effectiveness and efficiency compared to previous approaches.
Code will be available at \url{https://github.com/kugwzk/SmartTrim}.
 \\ \newline  }

\begin{document}

\maketitleabstract

\begin{figure}[t]
    \begin{center}
    \includegraphics[clip, width=\linewidth]{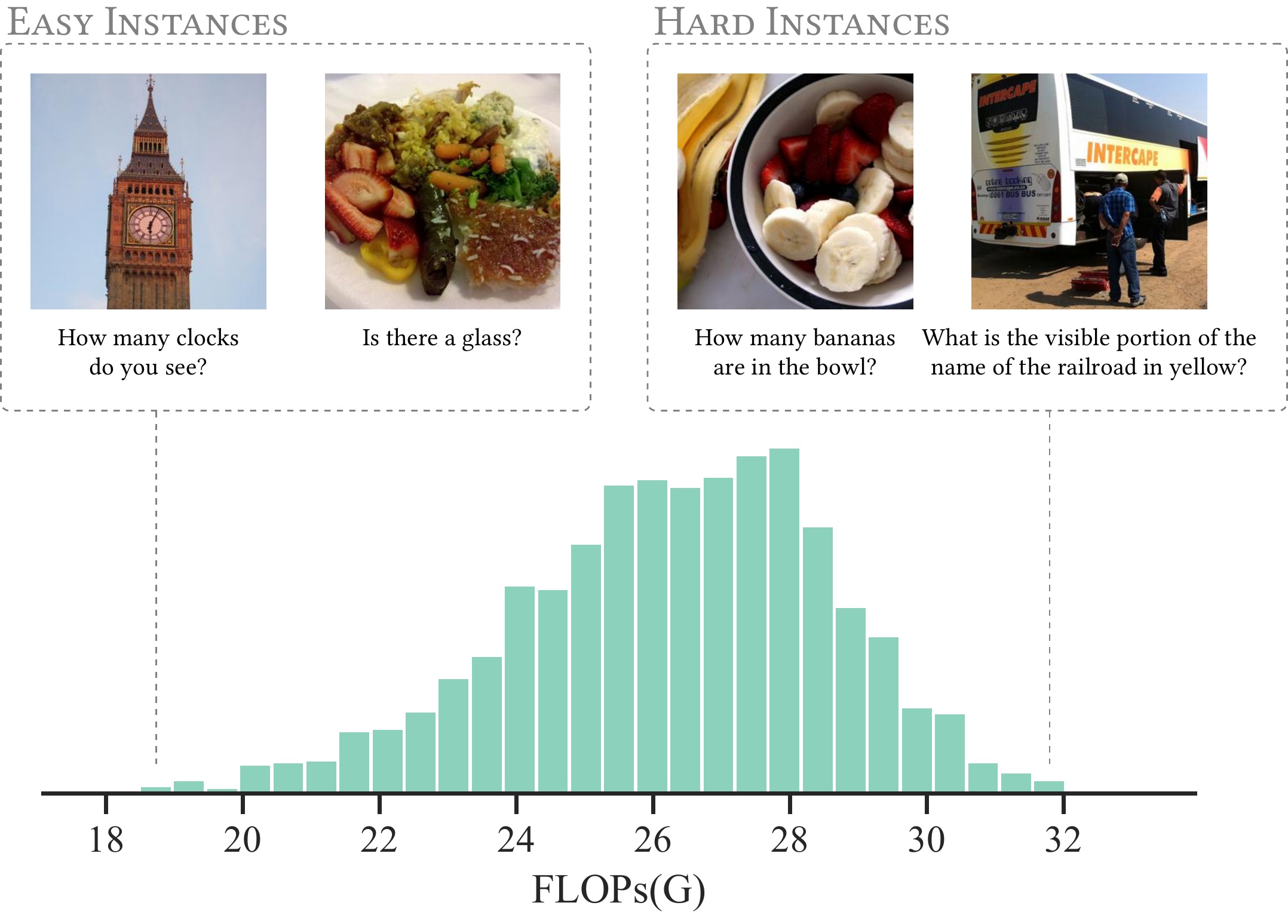}
    \caption{
    FLOPs histogram of \our{} on VQA.
    \our{} allocates diverse computational overhead based on cross-modal complexity, assigning fewer computations to \textbf{easy} instances (left) and more to \textbf{hard} ones (right).
    }
    \label{fig:least_most_mac}
    \end{center}
\end{figure}

\begin{figure*}[!t]
    \begin{center}
    \includegraphics[clip, width=\linewidth]{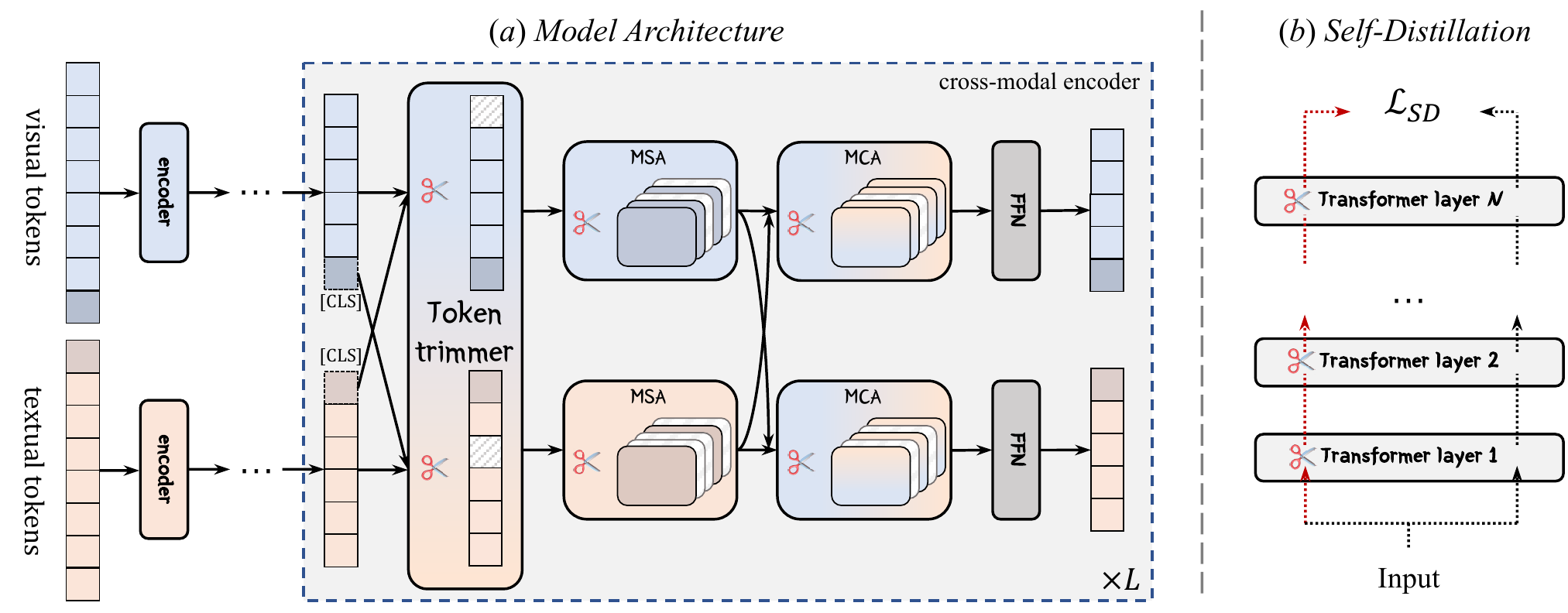}
    \caption{
    Overview of our \our{} framework, best viewed in color.
    (a) \textbf{Model Architecture} of \our{}.
    We incorporate the trimmers into layers of the uni-modal encoders and the cross-modal encoder to prune redundant tokens and heads.
    Given a set of image-text pairs, \our{} adjusts the computations for each instance based on the trimmer outputs.
    (b) \textbf{Self-Distillation} strategy.
    At each training step, the predictions of the pruned model are aligned to its fully-capacity counterpart.
    }
    \label{fig:overview}
    \end{center}
\end{figure*}

\section{Introduction}

Transformer-based~\citep{transformer} Vision-Language Models (VLMs) have shown great success on various vision-language tasks with their delicate model structures~\citep{clip, beit3, pali}.
Despite achieving superior performance, these models are computationally expensive due to the long input sequences and large number of parameters, hindering their deployment in the production environment.

In pursuit of efficient VLMs, a few acceleration approaches have been proposed, including knowledge distillation~\citep{DistillVLM, efficientvlm}, parameter pruning~\citep{DBLP:conf/aaai/GanCLC0WLW022, upop}, and token pruning~\citep{TRIPS,pumer}.
These methods reduce inference overhead, implying that a large proportion of parameters and token representations are redundant.
However, they adhere to a static computational architecture for all instances, overlooking the variation of complexities among different instances, leading to severe performance degradation at higher acceleration ratios~\citep{shallow_deep,fastbert}.
As demonstrated in Figure~\ref{fig:least_most_mac}, the instances involving complex cross-modal interactions naturally require more computations to fully comprehend the intricate details of images and associated questions.
Conversely, easy instances can be solved with less overhead.
Consequently, enormous original VLMs may overthink simple instances, leading to wasted computation, while static accelerated models struggle with complex ones, incurring extensive performance degradation.

To this end, we focus on adaptive acceleration on a per-input basis, which is orthogonal to static approaches and more flexible to meet different constraints.
In this work, we propose \our{}, an adaptive pruning framework for VLM (shown in Figure~\ref{fig:overview}), which streamlines the model from two aspects with significant redundancy: token representation and attention heads.
\our{} integrates the lightweight modules (called trimmers) into layers of the original backbone to identify redundant tokens and heads guided by cross-modal information.
Specifically, the \textit{XModal-aware token trimmers} are introduced to determine which tokens to retain considering not only their representations but also their importance in cross-modal interactions.
For head pruning, we introduce \textit{Modal-adaptive head trimmers} in different attention modules to adaptively select which heads to activate.
During training, we propose a self-distillation strategy, which encourages the predictions of the pruned model to align with its fully-capacity counterpart at the same step.
The self-distillation scheme alleviates the need for a separately fine-tuned teacher model in conventional knowledge distillation.
Furthermore, with a curriculum training scheduler, \our{} has a smoother and more stable optimization process.
Compared to previous methods, our approach not only avoids additional expensive pre-training, but also provides more fine-grained control to better explore efficiency-performance trade-offs.

We evaluate the proposed \our{} on two representative VLMs with different architectures: METER~\citep{meter}, an encoder-based model; and BLIP~\citep{blip}, an encoder-decoder-based model.
Experimental results reveal that \our{} consistently outperforms previous methods on various datasets.
Notably, \our{} achieves an impressive speed-up from $1.5\times$ to $4\times$ on the original model while incurring only a marginal performance drop ($1\%$\textasciitilde$3\%$).
Further analysis indicates that \our{} effectively learns to adaptively allocate computational budgets based on the complexity of cross-modal interactions.
\section{Preliminary}

\subsection{Transformer-based VLM}

\paragraph{Uni-Modal Encoders}
The input image and text are tokenized into visual and textual tokens, respectively.
The two sequences are fed into visual and textual encoders to extract the respective features, where each layer consists of a multi-head self-attention module (MSA) and a feed-forward network module (FFN).

\paragraph{Cross-Modal Encoder} 
To capture cross-modal interactions, the co-attention mechanism~\citep{vilbert} is employed in each layer of cross-modal encoder.
Specifically, in addition to MSA and FFN, a multi-head cross-attention module (MCA) is introduced, where query features are projected from one modality (\textit{e.g.}, vision), while key and value features are obtained from another modality (\textit{e.g.}, language).

\begin{figure}[t]
    \begin{center}
    \includegraphics[clip, width=\linewidth]{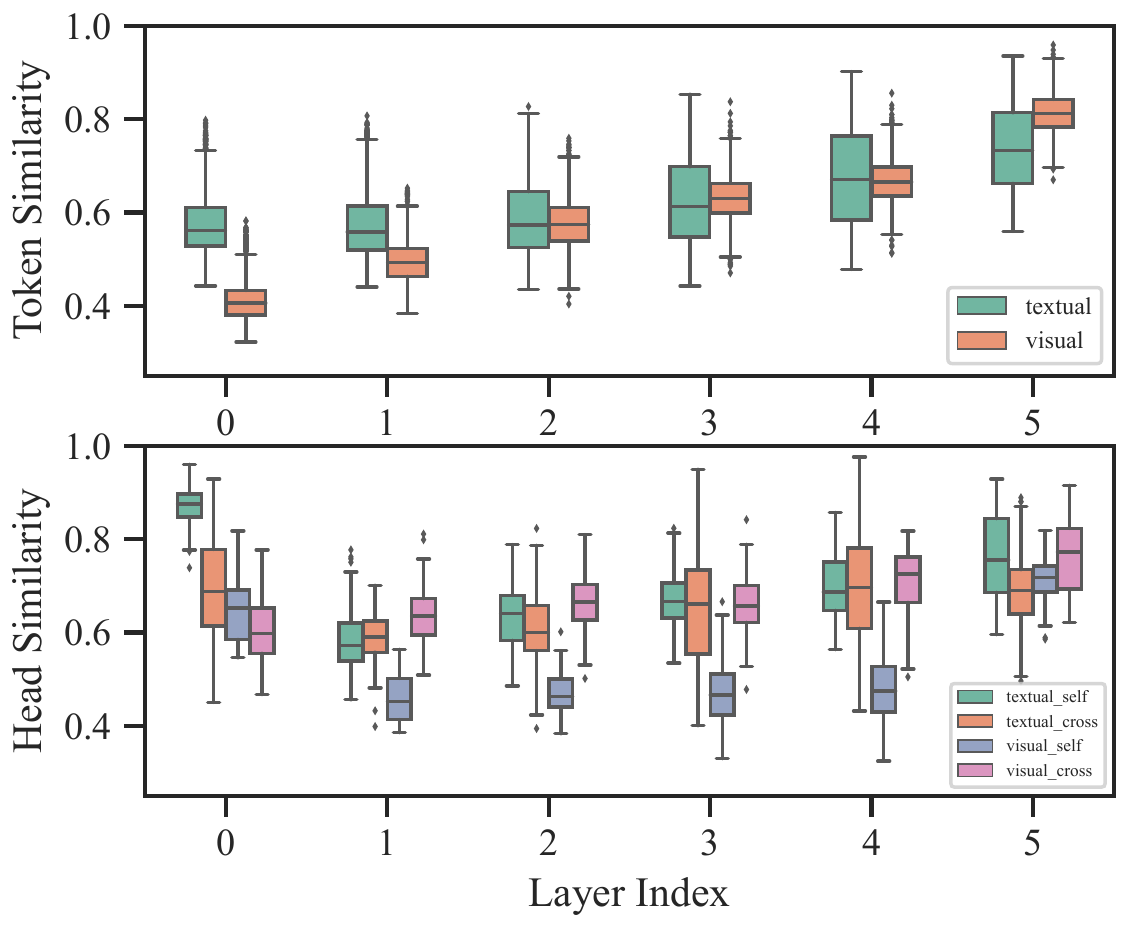}
    \caption{ 
        The similarities in representations of tokens (top) and heads  (bottom) in cross-modal encoder of METER fine-tuned on VQA.
    }
    \label{fig:vqa_cross_sim}
    \end{center}
\end{figure}

\subsection{Empirical Analyses}
\label{sec:redundancy_analysis}
The long sequence in VLMs incurs substantial computational overhead as the complexity of attention modules scales quadratically with length.
In addition, hundreds of millions of parameters further burden the situation.
Previous studies of uni-modal Transformers reveal that redundancy is present in token representations or attention heads~\citep{DBLP:conf/nips/MichelLN19, power_bert, redundancy_in_vit}.
To investigate whether redundancy also exists in VLMs, we measure cosine similarities between different token representations and heads at each layer of a fine-tuned METER.
As shown in Figure~\ref{fig:vqa_cross_sim}, our empirical findings are as follows:
\ding{182} Similarities between the representations of tokens and heads are consistently high across all layers, implying significant redundancy within the model.
\ding{183} The similarity of token representations increases progressively with depth, indicating a growing redundancy in deeper layers.
\ding{184} Similarities vary greatly between instances, prompting the need to investigate input-dependent adaptive pruning.
\section{Methodology}
In this section, we introduce the proposed adaptive pruning method for VLMs named \our{}, as shown in Figure~\ref{fig:overview}.
We first describe the details of adaptive trimmers and then introduce the end-to-end training recipe for \our{}.

\subsection{Adaptive Trimmers}
\paragraph{XModal-Aware Token Trimmer}
As shown in Figure~\ref{fig:overview} (a), \our{} progressively prunes token representations in blocks, delivering more important tokens to subsequent blocks, and eliminating the rest\footnote{We retain \sptk{CLS} tokens in each block of model.}.
To estimate the importance of token representations, we insert a lightweight MLP-based module (named \textit{XModal-aware trimmer}) before each block of uni-modal and cross-modal encoders.
Taking the cross-modal encoder block, for example, the $N_t$ token representations $\mX\in \R^{N_t\times D}$ are first fed into the \textit{local} policy network: 
\begin{equation*}
\bm{\pi^{l}_{t}} = \mathrm{MLP}_{t}(\mX^{\prime}) = \mathrm{MLP}_{t}(\mathrm{Linear}(\mX))
\end{equation*}
where $ \bm{\pi^{l}_{t}} \in \R^{N_t}$ is the local importance score of tokens, $\mX^{\prime}\in \R^{N_t\times D^{\prime}}$ is obtained by the dimension reduction of $\mX$.
The $\bm{\pi^{l}_{t}}$ is only computed based on the independent representations of tokens, without considering their contribution in cross-modal interactions.
To estimate the importance of cross-modal interactions without imposing excessive additional computation, we fuse global representations \footnote{We choose the representations of \sptk{CLS} tokens as global representations of each modality, which is better than other strategies in preliminary experiments, such as average or attentive pooling.} of visual and textual modality and then project to obtain the cross-modal global representation $\vg$, which contains global information of both modalities.
Then, we feed $\vg$ and $\mX^{\prime}$ to the \textit{global} policy network to calculate the XModal-global importance score $\bm{\pi_t^g}$:
\begin{equation*}
    \bm{\pi_t^{g}} = \mathrm{norm}(\vg \mW_g \mX^{\prime \intercal})
\end{equation*}
where $\mW_g$ is the projection layer.
The final token importance score $\bm{\pi_{t}}$ sums $\bm{\pi_t^{l}}$ and $\bm{\pi_t^{g}}$: $\pi_t = \pi_t^l + \pi_t^g$.
During inference, the pruning mask $\mM_{t} \in \{0, 1\}^{N_t}$ is sampled directly from $\mathrm{sigmoid}(\bm{\pi_{t}})$: $1$ indicates that the token is retained; otherwise, the token is removed.
By this pruning, our token trimmers reduce the amount of computation in both the attention and FFN modules for subsequent blocks.

\paragraph{Modal-adaptive Head Trimmer}
The VLMs capture intra-modal and inter-modal interactions via MSA and MCA, respectively.
However, the computational overhead required for modeling varies depending on the input complexity of attention, leading to redundancy in attention modules, as shown in Section~\ref{sec:redundancy_analysis}.
To this end, we integrate the \textit{modal-adaptive} head trimmer into the attention modules.
Specifically, we take the global representations of input sequences to feed into head trimmers:
\begin{equation*}
\bm{\pi_h} = \left\{
\begin{aligned}
    & \mathrm{MLP}_{h}^{self}(\vx_{\text{cls}}) & (\text{MSA}) \\
    & \mathrm{MLP}^{cross}_h([\vx_{\text{cls}}, \vy_{\text{cls}}])) & (\text{MCA})
\end{aligned}
\right.
\end{equation*}
where $\vx_{\text{cls}}, \vy_{\text{cls}}$ are the \sptk{CLS} representations of the self-modality and another modality, respectively.
Like the token trimmer, the head trimmer samples $\mM_{h}$ from $\mathrm{sigmoid}(\bm{\pi_h})$ to determine which heads to keep or remove.

Note that our trimmers introduce only a minor number of parameters ($3\%$) that yield a negligible computational overhead on FLOPs ($1\%$) compared to the original backbone.
In addition, adaptive trimmers are more hardware-friendly by avoiding the use of costly operations like top-\textit{k}in other methods~\citep{SpAtten}.

\subsection{Training Recipe}
The adaptive trimmers are seamlessly integrated into the backbone network fine-tuned with the task-specific objective $\mathcal{L}_{Task}$.
To achieve end-to-end optimization, we adopt the reparameterization technique~\citep{gumbel_softmax} to sample discrete masks $\mM$ from the output distributions of trimmers:
\begin{equation}
    \begin{split}
    \mM &= \frac{\exp((\bm{\pi} + \mG^{\prime}) / \tau)}{\exp((\bm{\pi} + \mG^{\prime}) / \tau) + \exp(\mG^{\prime\prime}/\tau)}\\
    \end{split}
    \label{eq:gumbel_sigmoid}
\end{equation}
where $\mG^{\prime}$ and $\mG^{\prime\prime}$ are two independent Gumbel noises, and $\tau$ is a temperature factor. 
To better control the overall computations of the model, we introduce a cost loss $\mathcal{L}_{Cost}$: 
\begin{align}
    \mathcal{L}_{Cost} = (\beta_{\mathcal{T}} - \gamma_{\mathcal{T}})^{2} + (\beta_{\mathcal{H}} - \gamma_{\mathcal{H}})^2 \\
    \beta_{\mathcal{T}} = \frac{1}{|\mathcal{T}|} \sum_{t \in \mathcal{T}} \frac{m_t}{N_{t}},
    \beta_{\mathcal{H}} = \frac{1}{|\mathcal{H}|} \sum_{h \in \mathcal{H}} \frac{m_h}{N_{h}}
\end{align}
where $\beta_\mathcal{T}$ and $\beta_\mathcal{H}$ represent the retention ratios of tokens and attention heads for each example in the batch. $\mathcal{T}$ and $\mathcal{H}$ are the sets of modules with token and head trimmers, respectively.
$\gamma$ is the overall target budget for token and head trimmers set in advance.
$m = \left\lVert \mM \right\rVert_0$ and $N$ represent the retained and total number of tokens or heads in the module.

\paragraph{Self-Distillation}
During training,
we propose a self-distillation objective to encourage the predictions of the pruned model $\theta_{s}$, to align with its fully-capacity counterpart $\theta_{t}$, as shown in Figure~\ref{fig:overview} (b).
Note that the $\theta_{s}$ and $\theta_{t}$ are \textbf{share} parameters, the only difference is that the trimmers are activated in the forward of $\theta_s$ while frozen in $\theta_t$.
At each training step, both the sparse and full models are optimized simultaneously.
The self-distillation objective $\mathcal{L}_{SD}$ is calculated as:
\begin{align*}
    \mathcal{L}_{SD} = \mathcal{L}_{Task}(\theta_{t}, y) + D_{\textit{KL}}(p(\theta_{s},x) \parallel p(\theta_{t},x))
\end{align*}
where $x$ is the input and $p$ are output logits.
This scheme alleviates the need for additional fine-tuned teacher models in traditional knowledge distillation.
The overall training objective of \our{} is as follows:
\begin{align}
    \mathcal{L} &= \mathcal{L}_{Task} + \lambda_{SD} \mathcal{L}_{SD} + \lambda_{Cost}\mathcal{L}_{Cost}
    \label{eq:overall_objective}
\end{align}
where $\lambda_{SD}, \lambda_{Cost}$ are hyperparameters.

\newcommand{\gray}{\color{gray}}

\begin{table*}[!t]
\centering
\small
\begin{tabular}{l|ccccccc|c}
\toprule
\multirow{2}{*}{\bf Methods} & \multicolumn{2}{c}{\bf NLVR2} & \bf VQA & \multicolumn{2}{c}{\bf SNLI-VE} & \multicolumn{2}{c|}{\bf ITR} & \multirow{2}{*}{\bf FLOPs(G)} \\
 & dev & test-P & test-dev & val & test & IR & TR &\\
\midrule
METER (backbone)~\citep{meter} & 82.05 & 82.32 & 77.43 & 81.24 & 80.91 & 92.5 & 98.1 & 88.5 \\ 
\midrule
\gray MiniVLM~\citep{minivlm} & \gray 73.71 & \gray 73.93 & \gray 69.10 & \gray - & \gray - & \gray - & \gray - & \gray - \\
\gray DistillVLM~\citep{DistillVLM} & \gray - & \gray - & \gray 69.80 & \gray - & \gray - & \gray - & \gray - & \gray - \\
\gray EfficientVLM~\citep{efficientvlm} & \gray 81.83 & \gray 81.72 & \gray 76.20 & \gray - & \gray - & \gray - & \gray - & \gray - \\
\midrule
\multicolumn{9}{l}{$\mathit{1.5\times}$ \textit{acceleration ratio}} \\
$\text{MuE}^{\dag}$~\citep{MuE} & 66.26 & 66.34 & 72.44 & 75.73 & 75.88 & 65.7 & 86.8 & 66.4 \\
$\text{TRIPS}^{\dag}$~\citep{TRIPS} & 81.34 & 82.01 & 76.50 & 80.55 & 80.57 & 91.8 & 97.5 & 59.0 \\
PuMer~\citep{pumer} & - & 82.20 & 76.80 & - & 80.30 & 91.7 & 97.6 & 64.7 \\
\our & \bf 81.89 & \bf 82.72 & \bf 77.25 & \bf 80.92 & \bf 80.90 & \bf 92.1 & \bf 97.9 & 56.0 \\
\midrule 
\multicolumn{9}{l}{$\mathit{2.0\times}$ \textit{acceleration ratio}} \\
FTKD & 76.89 & 77.49 & 68.23 & 77.12 & 77.21 & 77.1 & 86.5 & 48.2 \\
$\text{TRIPS}^{\dag}$~\citep{TRIPS} & 80.42 & 81.35 & 75.92 & 80.65 & 80.47 & 90.4 & 96.9 & 47.1 \\
\our & \bf 82.02 & \bf 81.97 & \bf 77.13 & \bf 80.67 & \bf 80.86 & \bf 91.6 & \bf 97.8 & 46.0 \\
\midrule
\multicolumn{9}{l}{$\mathit{2.5\times}$ \textit{acceleration ratio}} \\
FTKD & 65.86 & 67.10 & 59.32 & 73.30 & 73.27 & \xmark & \xmark & 32.4 \\
$\text{TRIPS}^{\dag}$~\citep{TRIPS} & 77.90 & 78.91 & 72.50 & 79.80 & 79.60 & 86.9 & 94.6 & 32.8 \\
\our & \bf 81.18 & \bf 81.55 & \bf 76.60 & \bf 80.53 & \bf 80.57 & \bf 89.8 & \bf 96.8 & 30.7 \\
\bottomrule
\end{tabular}

\caption{
    Results of acceleration methods on various downstream vision-language tasks with different acceleration ratios.
    FLOPs are measured on VQA with the same hyper-parameters.
    $\dag$ means the reimplementations by us.
    The marker \xmark~indicates methods do not achieve promising results.
    The best results for each ratio are marked with \textbf{boldface}.
    The results are averaged over 3 runs with different seeds.
    For a fair comparison, we de-emphasize MiniVLM, DistillVLM, EfficientVLM (by using gray color) since they require additional pre-training and based on different backbones.
}
\label{tab:main_meter}
\end{table*}

\let\gray\undefined

\begin{table*}[!t]
\centering
\small
\resizebox{\linewidth}{!}{
    \begin{tabular}{l|cccccccc}
    \toprule
    \multirow{2}{*}{\bf Methods} & \multicolumn{2}{c}{\bf NLVR2} & \multicolumn{1}{c}{\bf VQA} & \multicolumn{3}{c}{\bf COCO FT} & \multicolumn{2}{c}{\bf NoCaps ZS} \\
    & dev & test-P & test-dev & B@4 & C & S & C & S \\
    \midrule
    BLIP (backbone)~\citep{blip} & 82.57 & 82.53 & 78.2 & 39.9 & 133.3 & 23.8 & 109.3 & 14.7 \\
    \midrule
    \multicolumn{9}{l}{$\mathit{2.0\times}$ \textit{acceleration ratio}} \\
     UPop~\citep{upop} & 80.33 & 81.13 & 76.3 & - & 128.9 & 23.3 & - & - \\
     \our{} & \bf 82.24 & \bf 82.83 & \bf 78.0 & \bf 39.3 & \bf 130.8 & \bf 23.4 & \bf 106.4 & \bf 14.6 \\
     \midrule
    \multicolumn{9}{l}{$\mathit{4.0\times}$ \textit{acceleration ratio}} \\
     UPop~\citep{upop} & 72.85 & 73.55 & 74.5 & - & 117.4 & 21.7 & - & - \\
     \our{} & \bf 82.03 & \bf 82.35 & \bf 77.9 & \bf 38.2 & \bf 128.2 & \bf 23.0 & \bf 104.8 & \bf 14.2 \\
    \bottomrule
    \end{tabular}
}
\caption{
    Results of acceleration methods with BLIP backbone on various vision-language tasks across different acceleration ratios.
    The results are averaged over 3 runs with different seeds.
    B@4: BLEU@4, C: CIDEr, S: SPICE.
}
\label{tab:new_main_blip}
\end{table*}

\paragraph{Curriculum Training}
Integrating trimmers into the pretrained backbone introduces drastic adaptation to the original parameters, which potentially causes vulnerable and unstable training.
To enhance the stability of optimization, we propose a training scheduler driven by curriculum learning~\citep{curriculum_learning}.
Specifically, at the beginning of training, we initialize trimmers to ensure the retention of all tokens and heads.
Subsequently, we linearly decrease the ratio $\gamma$ from $1.0$ to the target ratio over a specified percentage of steps.
In this way, we encourage the training to focus on downstream tasks initially and then gradually learn adaptive pruning.
\section{Experiments}

\subsection{Setup}

\paragraph{Evaluation Datasets and Metrics}
We consider a diverse set of visual-language downstream tasks for evaluation: NLVR2~\citep{NLVR2}, VQA~\citep{vqav2} and SNLI-VE~\citep{snli_ve} for vision-language understanding, Flickr30K~\citep{Flickr} for image-text retrieval, COCO~\citep{coco_dataset} and NoCaps~\citep{NoCaps} for image captioning.
We report the accuracy for vision-language understanding tasks, and mean recall metrics for image retrieval (IR) and text retrieval (TR).
BLEU-4, CIDEr and SPICE are used to evaluate image captioning.

\paragraph{Implementation Details}
We adopt the pretrained METER and BLIP as backbones to initialize \our{}.
The adaptive trimmers consist of two linear layers with GeLU activation~\citep{gelu}, we set $D^{\prime} = D / 12$.
Fine-tuning hyperparameters mainly follow the defaults in \citet{meter} and \citet{blip}.
We set $\lambda_{\textit{Cost}}$ to $20.0$ and $\lambda_{\textit{SD}}$ to $1.0$.
Curriculum training is performed within the $60\%$ training step.
We employ FLOPs as the efficiency measurement of the models, which is hardware-independent\footnote{To prevent pseudo-improvement caused by pruning padding tokens, we evaluate without padding (single instance usage), similar to previous work~\citep{TR-BERT,adapLeR}.}.

\paragraph{Baselines}
We compare \our{} with the following VLM acceleration methods in the task-specific fine-tuning setting.
On the METER backbone:
\textbf{Fine-tuning Knowledge Distillation (FTKD)}, which initializes the student model by truncating the pretrained backbone following~\citet{pkd-bert} and then fine-tunes the model with logits/hidden representation/attention distillation objectives the same as \citet{tinybert}.
\textbf{TRIPS}~\citep{TRIPS}, which performs static token pruning based on attention scores to reduce the number of tokens in the visual encoder.
Note that we reimplement the method directly in the fine-tuning stage without additional pre-training for a fair comparison.
\textbf{PuMer}~\citep{pumer}, which is another static acceleration method that utilizes token pruning and merging.
Note that PuMer only prunes tokens in the cross-modal encoder.
\textbf{MuE}~\citep{MuE}, the only previous adaptive acceleration approach for VLM, which performs early exiting in terms of the similarities of layer-wise features.
We exhaustively search for the optimal settings and hyperparameters for the reimplemented baselines.
On the BLIP backbone, we mainly compare with the previous state-of-the-art method \textbf{UPop}~\citep{upop}, which simultaneously prunes and
retrains the backbone in a unified progressive pruning manner.
For reference, we also present the results of efficient VLMs that need additional pre-training, including MiniVLM~\citep{minivlm}, DistillVLM~\citep{DistillVLM} and EfficientVLM~\citep{efficientvlm}.

\subsection{Experimental Results}
\paragraph{Overall Performance}
We present the evaluation results based on the METER and BLIP architectures in Table~\ref{tab:main_meter} and Table~\ref{tab:new_main_blip}, respectively.
On the METER, \our{} effectively retains the performance of the original model ($97.1\%$ \textasciitilde~$100.0\%$), while enjoying considerable speed-up, ranging from $1.5\times$ to $2.5\times$.
To verify the generalizability of our approach, we also conduct an evaluation using BLIP as the backbone:
\our{} achieves competitive results compared to the original model in ratios of $2\times$ and $4\times$.
Compared to static acceleration baselines, \our{} significantly outperforms previous methods across various ratios and backbones, reflecting the effectiveness of our proposed adaptive pruning.
Furthermore, we observe that MuE, a previous adaptive acceleration VLM, performs poorly on challenging VL tasks (\textit{e.g.,} NLVR2 and VQA), which is due to its discarding of the entire layers of the model during inference.
In contrast, our \our{} focuses on more fine-grained units and delivers promising results even when applied at higher acceleration ratios.
In addition, \our{} achieves competitive performance compared to pretrained accelerated VLMs, further illustrating that our method is more economical.

\begin{figure}[!t]
    \begin{center}
    \includegraphics[clip, width=\linewidth]{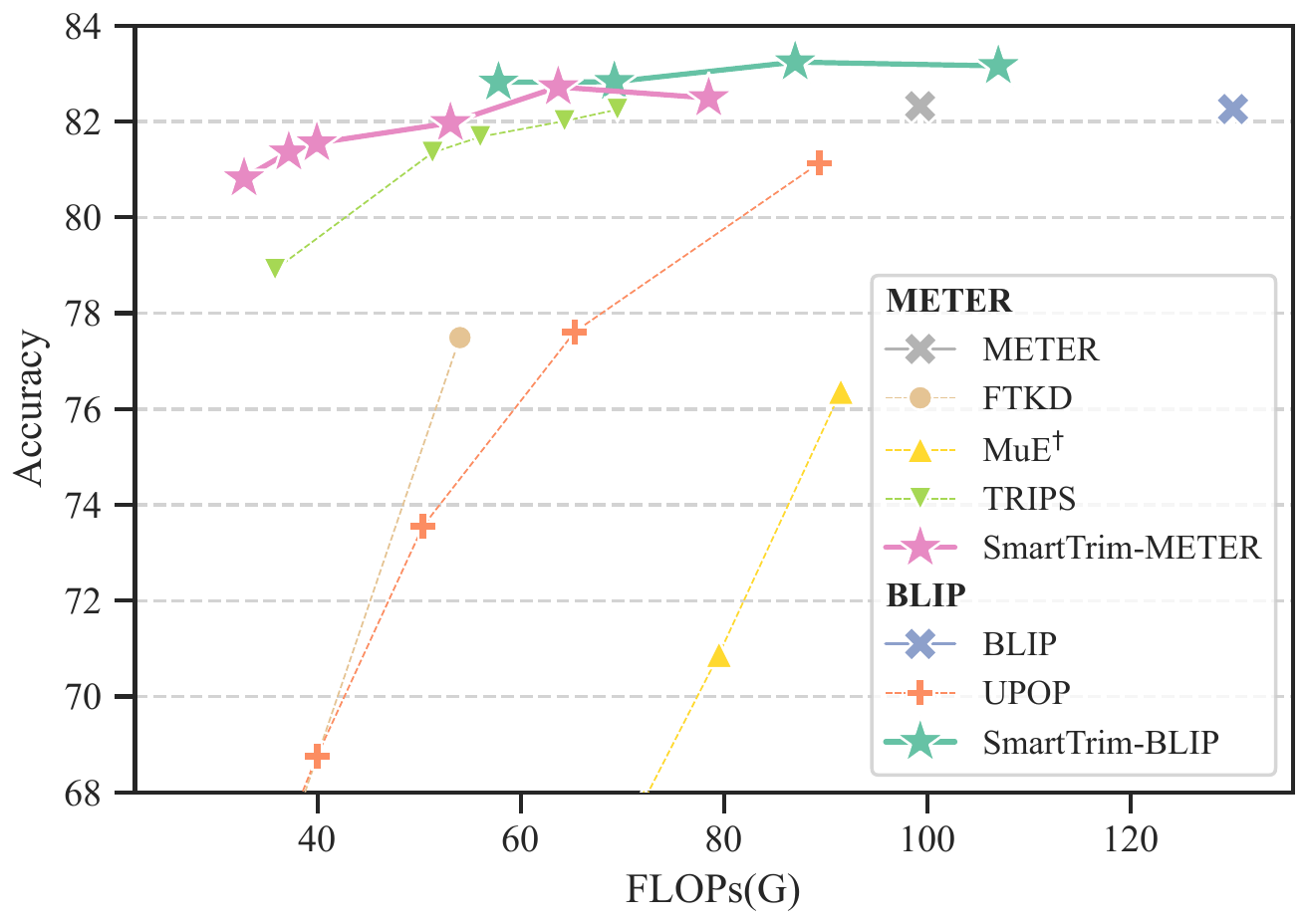}
    \caption{
    Pareto front of the efficiency-performance trade-offs of acceleration methods based on METER or BLIP backbones.
    }
    \label{fig:pareto}
    \end{center}
\end{figure}

\begin{figure}[!t]
    \begin{center}
    \includegraphics[clip, width=\linewidth]{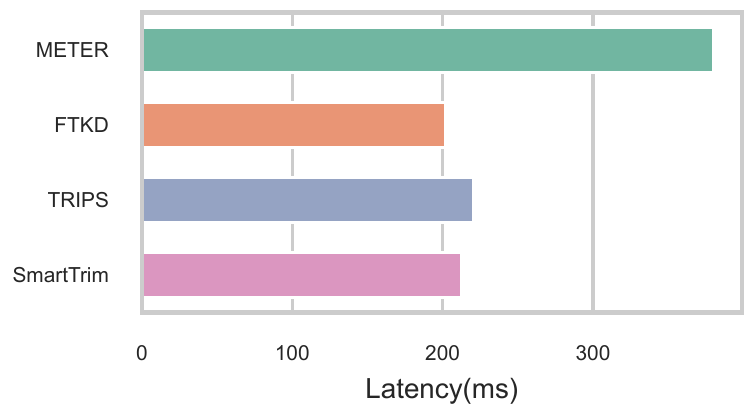}
    \caption{
    Averaged latency on the VQA dataset.
    }
    \label{fig:latency}
    \end{center}
\end{figure}

\paragraph{Efficiency-Performance Trade-offs}
Figure~\ref{fig:pareto} presents a Pareto front of efficiency-performance trade-offs of acceleration methods on NLVR2.
We observe that \our{} consistently outperforms other acceleration methods, especially at higher ratios (\textasciitilde$3.0\times$).
Surprisingly, \our{} performs even better than the original models with $21\%$\textasciitilde$35\%$ reduction in FLOPs, enjoying a "free lunch" in acceleration.
We further evaluate the latency of METER, FTKD, TRIPS, and \our{} on the VQA dataset.
The models are evaluated under the single-instance inference setting on the same CPU.
The results are shown in Figure~\ref{fig:latency}.
We find that \our{} is significantly faster than the original model.
Overall, \our{} achieves superior efficiency-performance trade-offs compared to the original models and previous acceleration methods.

\begin{table}[!t]
	\centering	
	\small
    \resizebox{\linewidth}{!}{
	\begin{tabular}	{l | c  c  c  c }
	\toprule	 	
	 \multirow{2}{*}{Models} & \multirow{2}{*}{Ratio} & \multicolumn{2}{c}{NLVR2} & VQA \\
	&  & dev & test-P & test-dev \\
	  \midrule
        \multirow{2}{*}{UPop}  & $2\times$ & 80.33 & 81.13 & 76.3 \\
        & $4\times$ & 72.85 & 73.55 & 74.5 \\
        \midrule
        $\text{UPop}_{2\times}+\our{}$ & $4\times$ & 80.52 & 80.85 & 76.0 \\
	\bottomrule
	\end{tabular}
 	}
	\caption{
            Results of adopting the static acceleration model UPop as the backbone.
            We also provide the target acceleration ratio for each model.
        }
	\label{tab:combine_upop}
	
\end{table}		
\newcommand{\floatingtext}[1]{\kern-0.1em\makebox[0pt][l]{#1}}
\newcommand{\up}[1]{\floatingtext{\color{deepgreen}\tiny #1}}
\newcommand{\down}[1]{\floatingtext{\color{red}\tiny #1}}

\begin{table}[!ht]
  \centering
  \small
  \begin{tabular}{l|ccc}
    \toprule
    \multirow{2}{*}{Models} & Image & VQA & \multirow{2}{*}{FLOPs(G)} \\
    & Resolution & test-dev & \\
    \midrule
    METER & $288^2$ & 76.78 & 48.3\\
    \our & $288^2$ & 76.44 & 26.2 \\
    \midrule
    METER & $384^2$ & 77.43 & 88.5 \\
    \our & $384^2$ & 77.13 & 46.0 \\
    \bottomrule
  \end{tabular}
  \caption{Results of models fine-tuned with different image resolutions on the VQA dataset.
  }
  \label{tab:high_resolution}
\end{table}

\let\bad\undefined
\let\up\undefined
\let\floatingtext\undefined

\paragraph{Combining with Static Acceleration Approaches}
The proposed \our{} is orthogonal to static acceleration approaches.
For further validation, we employ our approach on the static compressed model UPop, which statically prunes the parameters of the attention and FFN layers and achieves previous state-of-the-art performance on BLIP.
The training recipe for \our{} is easily augmented to UPop without changing the original fine-tuning process.
We utilize the UPop with the acceleration ratio $2\times$ as the backbone, and the results are presented in Table~\ref{tab:combine_upop}.
Comparing with $\text{UPop}_{2\times}$, we observe that \our{} can preserve over $99\%$ performance while enjoying faster inference.
This indicates that our adaptive pruning can effectively complement static acceleration approaches to achieve faster inference and smaller sizes for VLMs.
Moreover, \our{} significantly outperforms $\text{UPop}_{4\times}$, suggesting that combining \our{} with a static compression model may be better than directly training a smaller compression model, especially when aiming for higher speedup ratios.

\paragraph{Fine-tuning with different resolutions}
Table~\ref{tab:high_resolution} shows the VQA results of METER and \our{} on images of varying resolutions.
Our approach reduces the computational overhead of the original model, while maintaining performance on input images of different resolutions.
On METER models, increasing resolution improves results, but sacrifices efficiency, which poses a challenge in utilizing higher resolutions.
However, at higher resolution ($384^{2}$), \our{} retains performance while being even faster than METER with lower resolution ($288^{2}$), suggesting that \our{} can effectively encode images of higher resolution to improve performance while minimizing computational demands.

\section{Analysis}
\begin{figure}[t]
    \begin{center}
    \includegraphics[clip, width=\linewidth]{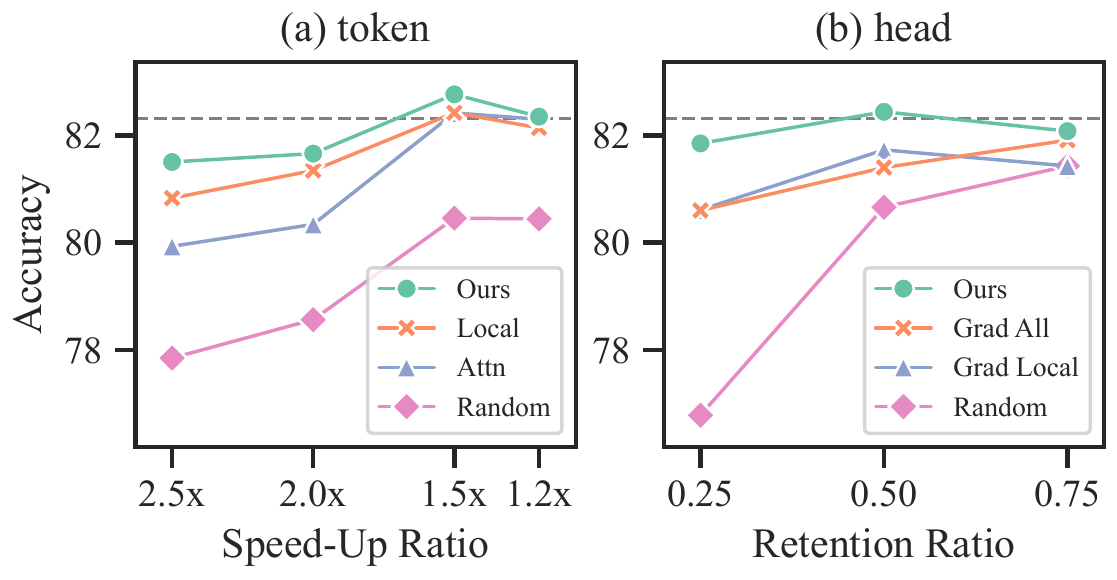}
    \caption{
    Comparison between different token (left) and head (right) pruning approaches on NLVR2.
    The dashed line denotes the performance of the original model.
    }
    \label{fig:token_head_compare}
    \end{center}
\end{figure}
In this section, we conduct extensive experiments to analyze \our{}.
All experiments are conducted on the METER backbone.

\subsection{Ablation Study}
\label{sec:ablation}
\paragraph{Effect of Adaptive Trimmers}
We first investigate the effect of our adaptive pruning trimmers.
For simplicity, we only consider the pruning in cross-modal encoder.
\ding{182} For \textit{token pruning}, we consider a variant of adaptive pruning without cross-modal guidance (\textit{Local}).
Besides, we also include static pruning baselines: random pruning (\textit{Random}) and attention score-based pruning (\textit{Attn};~\citet{TRIPS}).
We present the NLVR2 performance trend with different speed-up ratios in Figure~\ref{fig:token_head_compare}(a).
We find that both adaptive pruning methods outperform static pruning methods at various ratios.
Moreover, incorporating information from cross-modal interactions consistently improves performance, suggesting that cross-modal semantic guidance is critical to identifying more relevant tokens in different modalities.
\ding{183} For \textit{head pruning}, we compare with random pruning (\textit{Random}), and gradient-based pruning variants~\citep{DBLP:conf/nips/MichelLN19} including retaining top-$p$ heads in each module (\textit{Grad Local}) or in the whole model (\textit{Grad All}).
As shown in Figure~\ref{fig:token_head_compare}(b), our method significantly outperforms other baselines, especially in the low retention ratio regime ($0.25\times$), demonstrating the effectiveness of the proposed learned-based adaptive pruning mechanism.
Another interesting phenomenon is that a slight pruning of tokens and heads can improve performance, which can be seen as a “free lunch” of sparsity and also presented in BERT~\citep{attr} or ViT pruning~\citep{DBLP:conf/nips/ChenCGYZW21}.

\begin{table}[!t]
	\centering	
	\small
    \resizebox{\linewidth}{!}{
	\begin{tabular}	{l |  c  c  c }
	\toprule	 	
	 \multirow{2}{*}{Models} & \multicolumn{2}{c}{NLVR2} & VQA \\
	  & dev & test-P & test-dev \\
	  \midrule
        $\our{}_{1.5\times}$ & \bf 81.89 & \bf 82.72 & \bf 77.25 \\
        \quad - Self-Distillation & 81.58 & 82.50 & 77.06 \\
        \quad - Curriculum Training & 81.70 & 82.52 & 77.00 \\
        \midrule
        $\our{}_{2.0\times}$ & \bf 82.02 & 81.97 & \bf 77.13 \\
        \quad - Self-Distillation & 81.35 & 81.67 & 76.77 \\
        \quad - Curriculum Training & 81.58 & \bf 82.01 & 76.35 \\
        \midrule
        $\our{}_{2.5\times}$ & \bf 81.18 & \bf 81.55 & \bf 76.60 \\
        \quad - Self-Distillation & 80.51 & 81.30 & 75.79 \\
        \quad - Curriculum Training & 78.62 & 79.97 & 75.33 \\
	\bottomrule
	\end{tabular}
 	}
	\caption{
            Ablation studies of training strategies.
            Results are averaged over 3 runs.
        }
	\label{tab:training_ablation}
	
\end{table}		
\paragraph{Impact of Training Strategies}
We then analyze the impact of the proposed training strategies of \our{}.
As shown in Table~\ref{tab:training_ablation}, we compare the proposed \our{} with variants without self-distillation or curriculum training on the NLVR2 and VQA datasets.
From the results, we observe that both strategies improve performance at various acceleration ratios.
At higher acceleration ratios, these strategies make training more stable, leading to a dramatic improvement.

\subsection{Qualitative Analysis}
\label{sec:visualization}

\paragraph{Visualization of Token Trimming}

\begin{figure}[t]
    \begin{center}
    \includegraphics[clip, width=\linewidth]{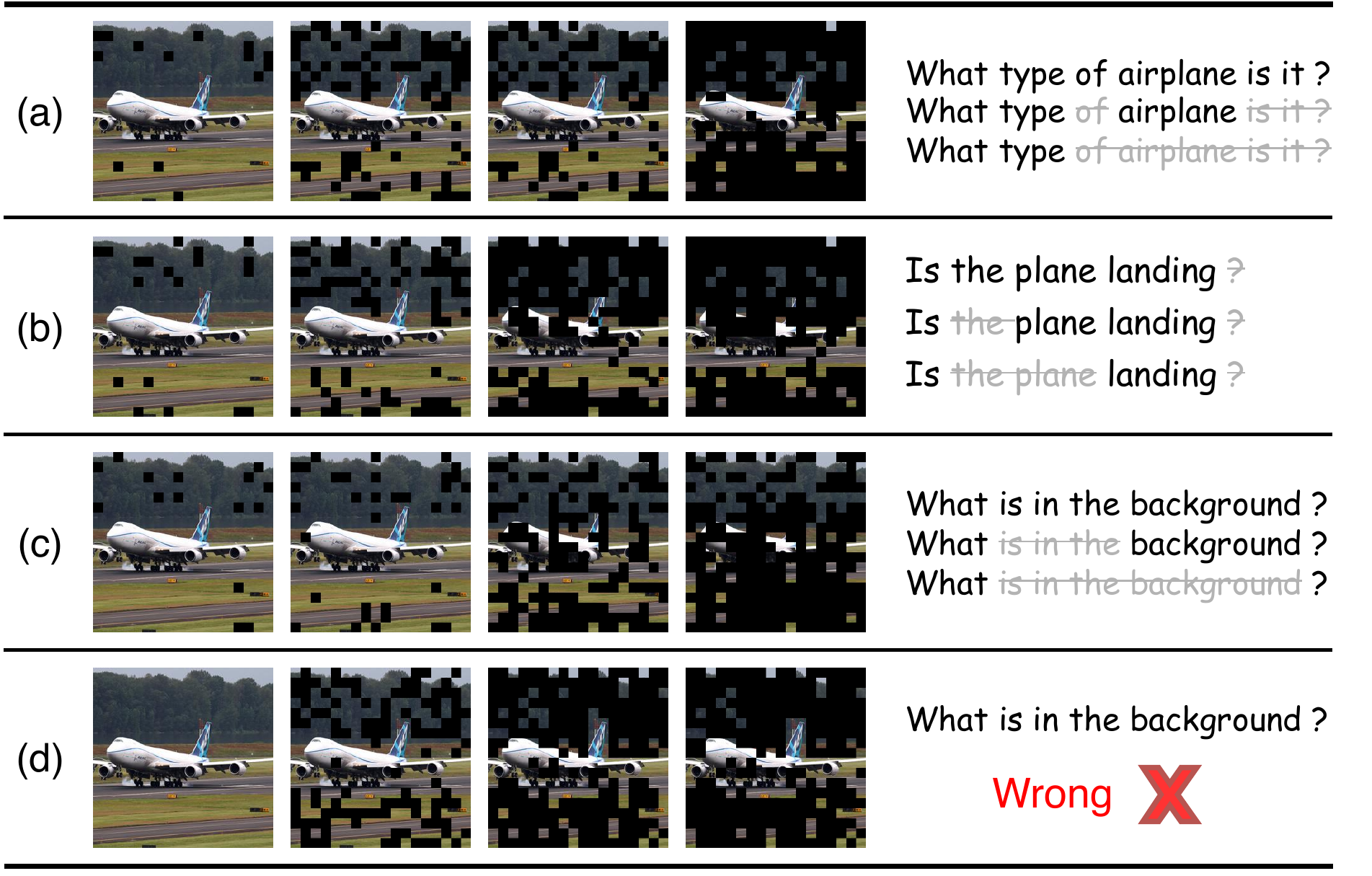}
    \caption{
    The visualizations of token trimming process on VQA.
    Image process order is shown from left to right and text is from top to bottom.
    (a)-(c) are obtained by our proposed XModal-aware token trimmer.
    (d) is from the local baseline that \textbf{without} cross-modal guidance, which finally yields a wrong answer.
    }
    \label{fig:token_visualization_compare}
    \end{center}
\end{figure}

We visualize the token trimming procedure in Figure~\ref{fig:token_visualization_compare}: (a)-(c) are from our XModel-aware token trimmer in \our{} while (d) is from the baseline without cross-modal guidance (\textit{Local}).
We observe that the XModal-aware trimmer gradually eliminates redundant tokens and finally focuses on informative ones.
With the same input image, it can effectively identify patches relevant to different questions, thereby giving correct answers.
However, the local baseline (Figure~\ref{fig:token_visualization_compare} (d)) only keeps the subject of the image (\textit{plane}) but is irrelevant to the questions.
See more results in Appendix~\ref{sec:more_visualization}.

\paragraph{Distribution of Retained Attention Heads}

\begin{figure}[t]
    \begin{center}
    \includegraphics[clip, width=\linewidth]{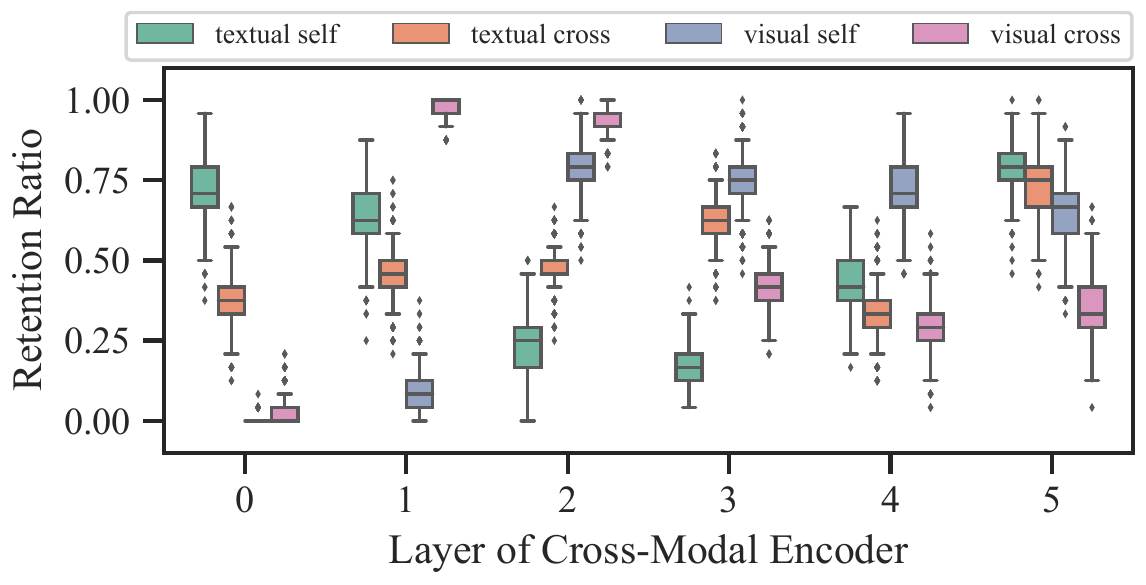}
    \caption{
     The head retention distribution of the model with $50\%$ target budget.
    }
    \label{fig:head_mask_boxplot}
    \end{center}
\end{figure}

Figure~\ref{fig:head_mask_boxplot} shows the distribution of the retention attention heads in \our{} with an overall target budget ratio of $50\%$.
We observe significant variations in retention heads between different instances, and \our{} learns distinct trimming strategies for different attention modules.

\paragraph{Adaptive Computational Patterns}
We further analyze the computational distribution of \our{} to investigate adaptive patterns.
We use a model with targeting on a $2$ times acceleration budget \footnote{The resolution of input images is $288^{2}$.} and show the visualization in Figure~\ref{fig:least_most_mac}.
As shown in Figure~\ref{fig:least_most_mac}, we observe that \our{} can achieve an acceleration ranging from $1.5\times$ to $2.7\times$ on various instances.
Furthermore, it learns to allocate more computations to instances that require complex cross-modal interactions and less to simple ones.
These findings indicate that \our{} can adaptively allocate computational overhead across diverse inputs.

\section{Related Work}
\subsection{Vision-Language Models}
The Transformer-based vision-language model (VLM) has emerged as a dominant architecture for various vision-language tasks~\citep{clip, vilt, albef, vlmo, ofa, coca, xvlm, bridgetower, blip2}.
Although they achieve satisfactory performance, the extensive amount of parameters inflicts an extravagant computational burden, impeding their scalability and application in the production environment.

\subsection{Transformer Acceleration}
Extensive research aims at accelerating Transformer, which can be categorized into two streams: \textit{Static} and \textit{Adaptive} approaches~\citep{xujingjing_survey}.

\paragraph{Static Approaches} yield accelerated models that remain static for all instances during inference after deployment.
Prior work effectively accelerates uni-modal Transformers through various techniques, such as knowledge distillation~\citep{hinton_kd, distillbert, pkd-bert, tinybert, BERT-of-Theseus, minilmv1}, parameter pruning~\citep{magnitude_pruning, DBLP:conf/nips/MichelLN19, DBLP:conf/emnlp/WangWL20, movement_pruning, dynabert, layerdropout, cofi}, and static token reduction via pruning~\citep{power_bert, DBLP:conf/nips/ChenCGYZW21, DynamicViT,DBLP:conf/cvpr/Tang00XGXT22,evit,evo_vit} or merging~\citep{tokenlearner, tome} less relevant tokens.
Recently, a few static methods dedicated to VLMs have been proposed~\citep{minivlm, dide, DistillVLM, DBLP:conf/aaai/GanCLC0WLW022}.
EfficientVLM~\citep{efficientvlm} is trained under a framework of pre-training distillation followed by pruning.
\citet{upop} introduces a progressive search-and-prune method, which needs retraining to sustain performance.
TRIPS~\citep{TRIPS} proposes to eliminate visual tokens using textual information by pre-training, while they only focus on token reduction in the visual encoder and keep trimming ratios static for all instances.
These methods require pre-training or iterative retraining to retain performance while being computationally expensive.
\citet{pumer} introduces static token pruning and merging within the VLM cross-modal encoder.
Overall, static acceleration fixes architecture regardless of large variations in the complexity of instances, limiting the capability of models.

\paragraph{Adaptive Approaches} enable accelerated models to adjust the computation required based on inputs dynamically.
Early exiting strategy has been applied to accelerate uni-modal Transformers by terminating inference at an early layer~\citep{DeeBert,pabee}.
Another stream is adaptive token pruning~\citep{TR-BERT,IA-RED2, ltp,Transkimmer,A-vit, adavit, spvit,efficientprompt}, which uses a policy network to gradually eliminate redundant tokens on a per-instance basis.
However, employing these uni-modal approaches directly in multimodal scenarios is suboptimal, as they overlook the importance of cross-modal interactions.
\citet{MuE} applies the early exiting technique based on layerwise similarities for an encoder-decoder-based VLM.
However, the constraint of pruning all tokens at the same layer is aggressive, resulting in significant performance degradation on challenge VL tasks, as shown in our experiments.
In contrast, \our{} focus on more fine-grained pruning units: token and attention heads, to achieve a better performance-efficiency trade-off.

\section{Conclusion}
In this work, we present~\our{}, an adaptive pruning framework for efficient VLMs that dynamically adjusts the computation overhead in an input-dependent manner.
By integrating token and head trimmers along with the backbone, \our{} prunes redundant tokens and heads during runtime based on the cross-modal information guidance and the pre-given budget.
Extensive experiments across various architectures and datasets show that \our{} achieves better efficiency-performance trade-offs.
We hope our endeavor will benefit end users by making multimodal systems more accessible. 


\section*{Acknowledgements}
We thank anonymous reviewers for their insightful feedback that helped improve the paper. 
The first two authors contributed equally.
The research is supported by the National Key Research and Development Project (2021YFF0901602), the National Science Foundation of China (U22B2059, 62276083), and Shenzhen Foundational Research Funding (JCYJ20200109113441941), Major Key Project of PCL (PCL2021A06).
Ming Liu is the corresponding author.

\section{Bibliographical References}\label{sec:reference}
\bibliographystyle{lrec-coling2024-natbib}
\bibliography{custom}

\begin{thebibliography}{72}
\expandafter\ifx\csname natexlab\endcsname\relax\def\natexlab#1{#1}\fi

\bibitem[{Agrawal et~al.(2019)Agrawal, Anderson, Desai, Wang, Chen, Jain, Johnson, Batra, Parikh, and Lee}]{NoCaps}
Harsh Agrawal, Peter Anderson, Karan Desai, Yufei Wang, Xinlei Chen, Rishabh Jain, Mark Johnson, Dhruv Batra, Devi Parikh, and Stefan Lee. 2019.
\newblock \href {https://doi.org/10.1109/ICCV.2019.00904} {nocaps: novel object captioning at scale}.
\newblock In \emph{2019 {IEEE/CVF} International Conference on Computer Vision, {ICCV} 2019, Seoul, Korea (South), October 27 - November 2, 2019}, pages 8947--8956. {IEEE}.

\bibitem[{Bao et~al.(2022)Bao, Wang, Dong, Liu, Mohammed, Aggarwal, Som, Piao, and Wei}]{vlmo}
Hangbo Bao, Wenhui Wang, Li~Dong, Qiang Liu, Owais~Khan Mohammed, Kriti Aggarwal, Subhojit Som, Songhao Piao, and Furu Wei. 2022.
\newblock \href {http://papers.nips.cc/paper\_files/paper/2022/hash/d46662aa53e78a62afd980a29e0c37ed-Abstract-Conference.html} {Vlmo: Unified vision-language pre-training with mixture-of-modality-experts}.
\newblock In \emph{Advances in Neural Information Processing Systems 35: Annual Conference on Neural Information Processing Systems 2022, NeurIPS 2022, New Orleans, LA, USA, November 28 - December 9, 2022}.

\bibitem[{Bengio et~al.(2009)Bengio, Louradour, Collobert, and Weston}]{curriculum_learning}
Yoshua Bengio, J{\'{e}}r{\^{o}}me Louradour, Ronan Collobert, and Jason Weston. 2009.
\newblock \href {https://doi.org/10.1145/1553374.1553380} {Curriculum learning}.
\newblock In \emph{Proceedings of the 26th Annual International Conference on Machine Learning, {ICML} 2009, Montreal, Quebec, Canada, June 14-18, 2009}, volume 382 of \emph{{ACM} International Conference Proceeding Series}, pages 41--48. {ACM}.

\bibitem[{Bolya et~al.(2023)Bolya, Fu, Dai, Zhang, Feichtenhofer, and Hoffman}]{tome}
Daniel Bolya, Cheng{-}Yang Fu, Xiaoliang Dai, Peizhao Zhang, Christoph Feichtenhofer, and Judy Hoffman. 2023.
\newblock \href {https://openreview.net/pdf?id=JroZRaRw7Eu} {Token merging: Your vit but faster}.
\newblock In \emph{The Eleventh International Conference on Learning Representations, {ICLR} 2023, Kigali, Rwanda, May 1-5, 2023}. OpenReview.net.

\bibitem[{Cao et~al.(2023)Cao, Paranjape, and Hajishirzi}]{pumer}
Qingqing Cao, Bhargavi Paranjape, and Hannaneh Hajishirzi. 2023.
\newblock \href {https://doi.org/10.18653/v1/2023.acl-long.721} {Pumer: Pruning and merging tokens for efficient vision language models}.
\newblock In \emph{Proceedings of the 61st Annual Meeting of the Association for Computational Linguistics (Volume 1: Long Papers), {ACL} 2023, Toronto, Canada, July 9-14, 2023}, pages 12890--12903. Association for Computational Linguistics.

\bibitem[{Chen et~al.(2021)Chen, Cheng, Gan, Yuan, Zhang, and Wang}]{DBLP:conf/nips/ChenCGYZW21}
Tianlong Chen, Yu~Cheng, Zhe Gan, Lu~Yuan, Lei Zhang, and Zhangyang Wang. 2021.
\newblock \href {https://proceedings.neurips.cc/paper/2021/hash/a61f27ab2165df0e18cc9433bd7f27c5-Abstract.html} {Chasing sparsity in vision transformers: An end-to-end exploration}.
\newblock In \emph{Advances in Neural Information Processing Systems 34: Annual Conference on Neural Information Processing Systems 2021, NeurIPS 2021, December 6-14, 2021, virtual}, pages 19974--19988.

\bibitem[{Chen et~al.(2023)Chen, Wang, Changpinyo, Piergiovanni, Padlewski, Salz, Goodman, Grycner, Mustafa, Beyer, Kolesnikov, Puigcerver, Ding, Rong, Akbari, Mishra, Xue, Thapliyal, Bradbury, and Kuo}]{pali}
Xi~Chen, Xiao Wang, Soravit Changpinyo, A.~J. Piergiovanni, Piotr Padlewski, Daniel Salz, Sebastian Goodman, Adam Grycner, Basil Mustafa, Lucas Beyer, Alexander Kolesnikov, Joan Puigcerver, Nan Ding, Keran Rong, Hassan Akbari, Gaurav Mishra, Linting Xue, Ashish~V. Thapliyal, James Bradbury, and Weicheng Kuo. 2023.
\newblock \href {https://openreview.net/pdf?id=mWVoBz4W0u} {Pali: {A} jointly-scaled multilingual language-image model}.
\newblock In \emph{The Eleventh International Conference on Learning Representations, {ICLR} 2023, Kigali, Rwanda, May 1-5, 2023}. OpenReview.net.

\bibitem[{Dou et~al.(2022)Dou, Xu, Gan, Wang, Wang, Wang, Zhu, Zhang, Yuan, Peng, Liu, and Zeng}]{meter}
Zi{-}Yi Dou, Yichong Xu, Zhe Gan, Jianfeng Wang, Shuohang Wang, Lijuan Wang, Chenguang Zhu, Pengchuan Zhang, Lu~Yuan, Nanyun Peng, Zicheng Liu, and Michael Zeng. 2022.
\newblock \href {https://doi.org/10.1109/CVPR52688.2022.01763} {An empirical study of training end-to-end vision-and-language transformers}.
\newblock In \emph{{IEEE/CVF} Conference on Computer Vision and Pattern Recognition, {CVPR} 2022, New Orleans, LA, USA, June 18-24, 2022}, pages 18145--18155. {IEEE}.

\bibitem[{Fan et~al.(2020)Fan, Grave, and Joulin}]{layerdropout}
Angela Fan, Edouard Grave, and Armand Joulin. 2020.
\newblock \href {https://openreview.net/forum?id=SylO2yStDr} {Reducing transformer depth on demand with structured dropout}.
\newblock In \emph{8th International Conference on Learning Representations, {ICLR} 2020, Addis Ababa, Ethiopia, April 26-30, 2020}. OpenReview.net.

\bibitem[{Fang et~al.(2021)Fang, Wang, Hu, Wang, Yang, and Liu}]{DistillVLM}
Zhiyuan Fang, Jianfeng Wang, Xiaowei Hu, Lijuan Wang, Yezhou Yang, and Zicheng Liu. 2021.
\newblock \href {https://doi.org/10.1109/ICCV48922.2021.00146} {Compressing visual-linguistic model via knowledge distillation}.
\newblock In \emph{2021 {IEEE/CVF} International Conference on Computer Vision, {ICCV} 2021, Montreal, QC, Canada, October 10-17, 2021}, pages 1408--1418. {IEEE}.

\bibitem[{Gan et~al.(2022)Gan, Chen, Li, Chen, Cheng, Wang, Liu, Wang, and Liu}]{DBLP:conf/aaai/GanCLC0WLW022}
Zhe Gan, Yen{-}Chun Chen, Linjie Li, Tianlong Chen, Yu~Cheng, Shuohang Wang, Jingjing Liu, Lijuan Wang, and Zicheng Liu. 2022.
\newblock \href {https://ojs.aaai.org/index.php/AAAI/article/view/19945} {Playing lottery tickets with vision and language}.
\newblock In \emph{Thirty-Sixth {AAAI} Conference on Artificial Intelligence, {AAAI} 2022, Thirty-Fourth Conference on Innovative Applications of Artificial Intelligence, {IAAI} 2022, The Twelveth Symposium on Educational Advances in Artificial Intelligence, {EAAI} 2022 Virtual Event, February 22 - March 1, 2022}, pages 652--660. {AAAI} Press.

\bibitem[{Goyal et~al.(2020)Goyal, Choudhury, Raje, Chakaravarthy, Sabharwal, and Verma}]{power_bert}
Saurabh Goyal, Anamitra~Roy Choudhury, Saurabh Raje, Venkatesan~T. Chakaravarthy, Yogish Sabharwal, and Ashish Verma. 2020.
\newblock \href {http://proceedings.mlr.press/v119/goyal20a.html} {Power-bert: Accelerating {BERT} inference via progressive word-vector elimination}.
\newblock In \emph{Proceedings of the 37th International Conference on Machine Learning, {ICML} 2020, 13-18 July 2020, Virtual Event}, volume 119 of \emph{Proceedings of Machine Learning Research}, pages 3690--3699. {PMLR}.

\bibitem[{Goyal et~al.(2017)Goyal, Khot, Summers{-}Stay, Batra, and Parikh}]{vqav2}
Yash Goyal, Tejas Khot, Douglas Summers{-}Stay, Dhruv Batra, and Devi Parikh. 2017.
\newblock \href {https://doi.org/10.1109/CVPR.2017.670} {Making the {V} in {VQA} matter: Elevating the role of image understanding in visual question answering}.
\newblock In \emph{2017 {IEEE} Conference on Computer Vision and Pattern Recognition, {CVPR} 2017, Honolulu, HI, USA, July 21-26, 2017}, pages 6325--6334. {IEEE} Computer Society.

\bibitem[{Guan et~al.(2022)Guan, Li, Leng, Lin, and Guo}]{Transkimmer}
Yue Guan, Zhengyi Li, Jingwen Leng, Zhouhan Lin, and Minyi Guo. 2022.
\newblock \href {https://doi.org/10.18653/v1/2022.acl-long.502} {Transkimmer: Transformer learns to layer-wise skim}.
\newblock In \emph{Proceedings of the 60th Annual Meeting of the Association for Computational Linguistics (Volume 1: Long Papers), {ACL} 2022, Dublin, Ireland, May 22-27, 2022}, pages 7275--7286. Association for Computational Linguistics.

\bibitem[{Han et~al.(2015)Han, Pool, Tran, and Dally}]{magnitude_pruning}
Song Han, Jeff Pool, John Tran, and William~J. Dally. 2015.
\newblock \href {https://proceedings.neurips.cc/paper/2015/hash/ae0eb3eed39d2bcef4622b2499a05fe6-Abstract.html} {Learning both weights and connections for efficient neural network}.
\newblock In \emph{Advances in Neural Information Processing Systems 28: Annual Conference on Neural Information Processing Systems 2015, December 7-12, 2015, Montreal, Quebec, Canada}, pages 1135--1143.

\bibitem[{Hao et~al.(2021)Hao, Dong, Wei, and Xu}]{attr}
Yaru Hao, Li~Dong, Furu Wei, and Ke~Xu. 2021.
\newblock \href {https://ojs.aaai.org/index.php/AAAI/article/view/17533} {Self-attention attribution: Interpreting information interactions inside transformer}.
\newblock In \emph{Thirty-Fifth {AAAI} Conference on Artificial Intelligence, {AAAI} 2021, Thirty-Third Conference on Innovative Applications of Artificial Intelligence, {IAAI} 2021, The Eleventh Symposium on Educational Advances in Artificial Intelligence, {EAAI} 2021, Virtual Event, February 2-9, 2021}, pages 12963--12971. {AAAI} Press.

\bibitem[{Hendrycks and Gimpel(2016)}]{gelu}
Dan Hendrycks and Kevin Gimpel. 2016.
\newblock \href {http://arxiv.org/abs/1606.08415} {Bridging nonlinearities and stochastic regularizers with gaussian error linear units}.
\newblock \emph{CoRR}, abs/1606.08415.

\bibitem[{Hinton et~al.(2015)Hinton, Vinyals, and Dean}]{hinton_kd}
Geoffrey~E. Hinton, Oriol Vinyals, and Jeffrey Dean. 2015.
\newblock \href {http://arxiv.org/abs/1503.02531} {Distilling the knowledge in a neural network}.
\newblock \emph{CoRR}, abs/1503.02531.

\bibitem[{Hou et~al.(2020)Hou, Huang, Shang, Jiang, Chen, and Liu}]{dynabert}
Lu~Hou, Zhiqi Huang, Lifeng Shang, Xin Jiang, Xiao Chen, and Qun Liu. 2020.
\newblock \href {https://proceedings.neurips.cc/paper/2020/hash/6f5216f8d89b086c18298e043bfe48ed-Abstract.html} {Dynabert: Dynamic {BERT} with adaptive width and depth}.
\newblock In \emph{Advances in Neural Information Processing Systems 33: Annual Conference on Neural Information Processing Systems 2020, NeurIPS 2020, December 6-12, 2020, virtual}.

\bibitem[{Jang et~al.(2017)Jang, Gu, and Poole}]{gumbel_softmax}
Eric Jang, Shixiang Gu, and Ben Poole. 2017.
\newblock \href {https://openreview.net/forum?id=rkE3y85ee} {Categorical reparameterization with gumbel-softmax}.
\newblock In \emph{5th International Conference on Learning Representations, {ICLR} 2017, Toulon, France, April 24-26, 2017, Conference Track Proceedings}. OpenReview.net.

\bibitem[{Jiang et~al.(2022)Jiang, Xu, Li, Yan, Ye, Zhang, Bi, and Huang}]{TRIPS}
Chaoya Jiang, Haiyang Xu, Chenliang Li, Ming Yan, Wei Ye, Shikun Zhang, Bin Bi, and Songfang Huang. 2022.
\newblock \href {https://aclanthology.org/2022.emnlp-main.273} {{TRIPS:} efficient vision-and-language pre-training with text-relevant image patch selection}.
\newblock In \emph{Proceedings of the 2022 Conference on Empirical Methods in Natural Language Processing, {EMNLP} 2022, Abu Dhabi, United Arab Emirates, December 7-11, 2022}, pages 4084--4096. Association for Computational Linguistics.

\bibitem[{Jiao et~al.(2020)Jiao, Yin, Shang, Jiang, Chen, Li, Wang, and Liu}]{tinybert}
Xiaoqi Jiao, Yichun Yin, Lifeng Shang, Xin Jiang, Xiao Chen, Linlin Li, Fang Wang, and Qun Liu. 2020.
\newblock \href {https://doi.org/10.18653/v1/2020.findings-emnlp.372} {Tinybert: Distilling {BERT} for natural language understanding}.
\newblock In \emph{Findings of the Association for Computational Linguistics: {EMNLP} 2020, Online Event, 16-20 November 2020}, volume {EMNLP} 2020 of \emph{Findings of {ACL}}, pages 4163--4174. Association for Computational Linguistics.

\bibitem[{Karpathy and Fei{-}Fei(2015)}]{Karpathy_split}
Andrej Karpathy and Li~Fei{-}Fei. 2015.
\newblock \href {https://doi.org/10.1109/CVPR.2015.7298932} {Deep visual-semantic alignments for generating image descriptions}.
\newblock In \emph{{IEEE} Conference on Computer Vision and Pattern Recognition, {CVPR} 2015, Boston, MA, USA, June 7-12, 2015}, pages 3128--3137. {IEEE} Computer Society.

\bibitem[{Kaya et~al.(2019)Kaya, Hong, and Dumitras}]{shallow_deep}
Yigitcan Kaya, Sanghyun Hong, and Tudor Dumitras. 2019.
\newblock \href {http://proceedings.mlr.press/v97/kaya19a.html} {Shallow-deep networks: Understanding and mitigating network overthinking}.
\newblock In \emph{Proceedings of the 36th International Conference on Machine Learning, {ICML} 2019, 9-15 June 2019, Long Beach, California, {USA}}, volume~97 of \emph{Proceedings of Machine Learning Research}, pages 3301--3310. {PMLR}.

\bibitem[{Kim et~al.(2022)Kim, Shen, Thorsley, Gholami, Kwon, Hassoun, and Keutzer}]{ltp}
Sehoon Kim, Sheng Shen, David Thorsley, Amir Gholami, Woosuk Kwon, Joseph Hassoun, and Kurt Keutzer. 2022.
\newblock \href {https://doi.org/10.1145/3534678.3539260} {Learned token pruning for transformers}.
\newblock In \emph{{KDD} '22: The 28th {ACM} {SIGKDD} Conference on Knowledge Discovery and Data Mining, Washington, DC, USA, August 14 - 18, 2022}, pages 784--794. {ACM}.

\bibitem[{Kim et~al.(2021)Kim, Son, and Kim}]{vilt}
Wonjae Kim, Bokyung Son, and Ildoo Kim. 2021.
\newblock \href {http://proceedings.mlr.press/v139/kim21k.html} {Vilt: Vision-and-language transformer without convolution or region supervision}.
\newblock In \emph{Proceedings of the 38th International Conference on Machine Learning}, volume 139 of \emph{Proceedings of Machine Learning Research}, pages 5583--5594. PMLR.

\bibitem[{Kong et~al.(2022)Kong, Dong, Ma, Meng, Niu, Sun, Shen, Yuan, Ren, Tang, Qin, and Wang}]{spvit}
Zhenglun Kong, Peiyan Dong, Xiaolong Ma, Xin Meng, Wei Niu, Mengshu Sun, Xuan Shen, Geng Yuan, Bin Ren, Hao Tang, Minghai Qin, and Yanzhi Wang. 2022.
\newblock \href {https://doi.org/10.1007/978-3-031-20083-0\_37} {Spvit: Enabling faster vision transformers via latency-aware soft token pruning}.
\newblock In \emph{Computer Vision - {ECCV} 2022 - 17th European Conference, Tel Aviv, Israel, October 23-27, 2022, Proceedings, Part {XI}}, volume 13671 of \emph{Lecture Notes in Computer Science}, pages 620--640. Springer.

\bibitem[{Li et~al.(2023)Li, Li, Savarese, and Hoi}]{blip2}
Junnan Li, Dongxu Li, Silvio Savarese, and Steven C.~H. Hoi. 2023.
\newblock \href {https://proceedings.mlr.press/v202/li23q.html} {{BLIP-2:} bootstrapping language-image pre-training with frozen image encoders and large language models}.
\newblock In \emph{International Conference on Machine Learning, {ICML} 2023, 23-29 July 2023, Honolulu, Hawaii, {USA}}, volume 202 of \emph{Proceedings of Machine Learning Research}, pages 19730--19742. {PMLR}.

\bibitem[{Li et~al.(2022)Li, Li, Xiong, and Hoi}]{blip}
Junnan Li, Dongxu Li, Caiming Xiong, and Steven C.~H. Hoi. 2022.
\newblock \href {https://proceedings.mlr.press/v162/li22n.html} {{BLIP:} bootstrapping language-image pre-training for unified vision-language understanding and generation}.
\newblock In \emph{International Conference on Machine Learning, {ICML} 2022, 17-23 July 2022, Baltimore, Maryland, {USA}}, volume 162 of \emph{Proceedings of Machine Learning Research}, pages 12888--12900. {PMLR}.

\bibitem[{Li et~al.(2021)Li, Selvaraju, Gotmare, Joty, Xiong, and Hoi}]{albef}
Junnan Li, Ramprasaath~R. Selvaraju, Akhilesh Gotmare, Shafiq~R. Joty, Caiming Xiong, and Steven~Chu{-}Hong Hoi. 2021.
\newblock \href {https://proceedings.neurips.cc/paper/2021/hash/505259756244493872b7709a8a01b536-Abstract.html} {Align before fuse: Vision and language representation learning with momentum distillation}.
\newblock In \emph{Advances in Neural Information Processing Systems 34: Annual Conference on Neural Information Processing Systems 2021, NeurIPS 2021, December 6-14, 2021, virtual}, pages 9694--9705.

\bibitem[{Liang et~al.(2022)Liang, Ge, Tong, Song, Wang, and Xie}]{evit}
Youwei Liang, Chongjian Ge, Zhan Tong, Yibing Song, Jue Wang, and Pengtao Xie. 2022.
\newblock \href {https://openreview.net/forum?id=BjyvwnXXVn\_} {Evit: Expediting vision transformers via token reorganizations}.
\newblock In \emph{The Tenth International Conference on Learning Representations, {ICLR} 2022, Virtual Event, April 25-29, 2022}. OpenReview.net.

\bibitem[{Lin et~al.(2014)Lin, Maire, Belongie, Hays, Perona, Ramanan, Doll{\'{a}}r, and Zitnick}]{coco_dataset}
Tsung{-}Yi Lin, Michael Maire, Serge~J. Belongie, James Hays, Pietro Perona, Deva Ramanan, Piotr Doll{\'{a}}r, and C.~Lawrence Zitnick. 2014.
\newblock \href {https://doi.org/10.1007/978-3-319-10602-1\_48} {Microsoft {COCO:} common objects in context}.
\newblock In \emph{Computer Vision - {ECCV} 2014 - 13th European Conference, Zurich, Switzerland, September 6-12, 2014, Proceedings, Part {V}}, volume 8693 of \emph{Lecture Notes in Computer Science}, pages 740--755. Springer.

\bibitem[{Liu et~al.(2020)Liu, Zhou, Wang, Zhao, Deng, and Ju}]{fastbert}
Weijie Liu, Peng Zhou, Zhiruo Wang, Zhe Zhao, Haotang Deng, and Qi~Ju. 2020.
\newblock \href {https://doi.org/10.18653/v1/2020.acl-main.537} {Fastbert: a self-distilling {BERT} with adaptive inference time}.
\newblock In \emph{Proceedings of the 58th Annual Meeting of the Association for Computational Linguistics, {ACL} 2020, Online, July 5-10, 2020}, pages 6035--6044. Association for Computational Linguistics.

\bibitem[{Lu et~al.(2019)Lu, Batra, Parikh, and Lee}]{vilbert}
Jiasen Lu, Dhruv Batra, Devi Parikh, and Stefan Lee. 2019.
\newblock \href {https://proceedings.neurips.cc/paper/2019/hash/c74d97b01eae257e44aa9d5bade97baf-Abstract.html} {Vilbert: Pretraining task-agnostic visiolinguistic representations for vision-and-language tasks}.
\newblock In \emph{Advances in Neural Information Processing Systems 32: Annual Conference on Neural Information Processing Systems 2019, NeurIPS 2019, December 8-14, 2019, Vancouver, BC, Canada}, pages 13--23.

\bibitem[{Meng et~al.(2022)Meng, Li, Chen, Lan, Wu, Jiang, and Lim}]{adavit}
Lingchen Meng, Hengduo Li, Bor{-}Chun Chen, Shiyi Lan, Zuxuan Wu, Yu{-}Gang Jiang, and Ser{-}Nam Lim. 2022.
\newblock \href {https://doi.org/10.1109/CVPR52688.2022.01199} {Adavit: Adaptive vision transformers for efficient image recognition}.
\newblock In \emph{{IEEE/CVF} Conference on Computer Vision and Pattern Recognition, {CVPR} 2022, New Orleans, LA, USA, June 18-24, 2022}, pages 12299--12308. {IEEE}.

\bibitem[{Michel et~al.(2019)Michel, Levy, and Neubig}]{DBLP:conf/nips/MichelLN19}
Paul Michel, Omer Levy, and Graham Neubig. 2019.
\newblock \href {https://proceedings.neurips.cc/paper/2019/hash/2c601ad9d2ff9bc8b282670cdd54f69f-Abstract.html} {Are sixteen heads really better than one?}
\newblock In \emph{Advances in Neural Information Processing Systems 32: Annual Conference on Neural Information Processing Systems 2019, NeurIPS 2019, December 8-14, 2019, Vancouver, BC, Canada}, pages 14014--14024.

\bibitem[{Modarressi et~al.(2022)Modarressi, Mohebbi, and Pilehvar}]{adapLeR}
Ali Modarressi, Hosein Mohebbi, and Mohammad~Taher Pilehvar. 2022.
\newblock \href {https://doi.org/10.18653/v1/2022.acl-long.1} {Adapler: Speeding up inference by adaptive length reduction}.
\newblock In \emph{Proceedings of the 60th Annual Meeting of the Association for Computational Linguistics (Volume 1: Long Papers), {ACL} 2022, Dublin, Ireland, May 22-27, 2022}, pages 1--15. Association for Computational Linguistics.

\bibitem[{Pan et~al.(2021)Pan, Panda, Jiang, Wang, Feris, and Oliva}]{IA-RED2}
Bowen Pan, Rameswar Panda, Yifan Jiang, Zhangyang Wang, Rog{\'{e}}rio Feris, and Aude Oliva. 2021.
\newblock \href {https://proceedings.neurips.cc/paper/2021/hash/d072677d210ac4c03ba046120f0802ec-Abstract.html} {Ia-red{\textdollar}{\^{}}2{\textdollar}: Interpretability-aware redundancy reduction for vision transformers}.
\newblock In \emph{Advances in Neural Information Processing Systems 34: Annual Conference on Neural Information Processing Systems 2021, NeurIPS 2021, December 6-14, 2021, virtual}, pages 24898--24911.

\bibitem[{Plummer et~al.(2015)Plummer, Wang, Cervantes, Caicedo, Hockenmaier, and Lazebnik}]{Flickr}
Bryan~A. Plummer, Liwei Wang, Chris~M. Cervantes, Juan~C. Caicedo, Julia Hockenmaier, and Svetlana Lazebnik. 2015.
\newblock \href {https://doi.org/10.1109/ICCV.2015.303} {Flickr30k entities: Collecting region-to-phrase correspondences for richer image-to-sentence models}.
\newblock In \emph{2015 {IEEE} International Conference on Computer Vision, {ICCV} 2015, Santiago, Chile, December 7-13, 2015}, pages 2641--2649. {IEEE} Computer Society.

\bibitem[{Radford et~al.(2021)Radford, Kim, Hallacy, Ramesh, Goh, Agarwal, Sastry, Askell, Mishkin, Clark, Krueger, and Sutskever}]{clip}
Alec Radford, Jong~Wook Kim, Chris Hallacy, Aditya Ramesh, Gabriel Goh, Sandhini Agarwal, Girish Sastry, Amanda Askell, Pamela Mishkin, Jack Clark, Gretchen Krueger, and Ilya Sutskever. 2021.
\newblock \href {http://proceedings.mlr.press/v139/radford21a.html} {Learning transferable visual models from natural language supervision}.
\newblock In \emph{Proceedings of the 38th International Conference on Machine Learning, {ICML} 2021, 18-24 July 2021, Virtual Event}, volume 139 of \emph{Proceedings of Machine Learning Research}, pages 8748--8763. {PMLR}.

\bibitem[{Rao et~al.(2021)Rao, Zhao, Liu, Lu, Zhou, and Hsieh}]{DynamicViT}
Yongming Rao, Wenliang Zhao, Benlin Liu, Jiwen Lu, Jie Zhou, and Cho{-}Jui Hsieh. 2021.
\newblock \href {https://proceedings.neurips.cc/paper/2021/hash/747d3443e319a22747fbb873e8b2f9f2-Abstract.html} {Dynamicvit: Efficient vision transformers with dynamic token sparsification}.
\newblock In \emph{Advances in Neural Information Processing Systems 34: Annual Conference on Neural Information Processing Systems 2021, NeurIPS 2021, December 6-14, 2021, virtual}, pages 13937--13949.

\bibitem[{Ryoo et~al.(2021)Ryoo, Piergiovanni, Arnab, Dehghani, and Angelova}]{tokenlearner}
Michael~S. Ryoo, A.~J. Piergiovanni, Anurag Arnab, Mostafa Dehghani, and Anelia Angelova. 2021.
\newblock \href {https://proceedings.neurips.cc/paper/2021/hash/6a30e32e56fce5cf381895dfe6ca7b6f-Abstract.html} {Tokenlearner: Adaptive space-time tokenization for videos}.
\newblock In \emph{Advances in Neural Information Processing Systems 34: Annual Conference on Neural Information Processing Systems 2021, NeurIPS 2021, December 6-14, 2021, virtual}, pages 12786--12797.

\bibitem[{Sanh et~al.(2019)Sanh, Debut, Chaumond, and Wolf}]{distillbert}
Victor Sanh, Lysandre Debut, Julien Chaumond, and Thomas Wolf. 2019.
\newblock \href {http://arxiv.org/abs/1910.01108} {Distilbert, a distilled version of {BERT:} smaller, faster, cheaper and lighter}.
\newblock \emph{CoRR}, abs/1910.01108.

\bibitem[{Sanh et~al.(2020)Sanh, Wolf, and Rush}]{movement_pruning}
Victor Sanh, Thomas Wolf, and Alexander~M. Rush. 2020.
\newblock \href {https://proceedings.neurips.cc/paper/2020/hash/eae15aabaa768ae4a5993a8a4f4fa6e4-Abstract.html} {Movement pruning: Adaptive sparsity by fine-tuning}.
\newblock In \emph{Advances in Neural Information Processing Systems 33: Annual Conference on Neural Information Processing Systems 2020, NeurIPS 2020, December 6-12, 2020, virtual}.

\bibitem[{Shi et~al.(2023)Shi, Tao, Jin, Yang, Yuan, and Wang}]{upop}
Dachuan Shi, Chaofan Tao, Ying Jin, Zhendong Yang, Chun Yuan, and Jiaqi Wang. 2023.
\newblock \href {https://proceedings.mlr.press/v202/shi23e.html} {Upop: Unified and progressive pruning for compressing vision-language transformers}.
\newblock In \emph{International Conference on Machine Learning, {ICML} 2023, 23-29 July 2023, Honolulu, Hawaii, {USA}}, volume 202 of \emph{Proceedings of Machine Learning Research}, pages 31292--31311. {PMLR}.

\bibitem[{Suhr et~al.(2019)Suhr, Zhou, Zhang, Zhang, Bai, and Artzi}]{NLVR2}
Alane Suhr, Stephanie Zhou, Ally Zhang, Iris Zhang, Huajun Bai, and Yoav Artzi. 2019.
\newblock \href {https://doi.org/10.18653/v1/p19-1644} {A corpus for reasoning about natural language grounded in photographs}.
\newblock In \emph{Proceedings of the 57th Conference of the Association for Computational Linguistics, {ACL} 2019, Florence, Italy, July 28- August 2, 2019, Volume 1: Long Papers}, pages 6418--6428. Association for Computational Linguistics.

\bibitem[{Sun et~al.(2019)Sun, Cheng, Gan, and Liu}]{pkd-bert}
Siqi Sun, Yu~Cheng, Zhe Gan, and Jingjing Liu. 2019.
\newblock \href {https://doi.org/10.18653/v1/D19-1441} {Patient knowledge distillation for {BERT} model compression}.
\newblock In \emph{Proceedings of the 2019 Conference on Empirical Methods in Natural Language Processing and the 9th International Joint Conference on Natural Language Processing, {EMNLP-IJCNLP} 2019, Hong Kong, China, November 3-7, 2019}, pages 4322--4331. Association for Computational Linguistics.

\bibitem[{Tang et~al.(2023)Tang, Wang, Kong, Zhang, Li, Ding, Wang, Liang, and Xu}]{MuE}
Shengkun Tang, Yaqing Wang, Zhenglun Kong, Tianchi Zhang, Yao Li, Caiwen Ding, Yanzhi Wang, Yi~Liang, and Dongkuan Xu. 2023.
\newblock \href {https://doi.org/10.1109/CVPR52729.2023.01038} {You need multiple exiting: Dynamic early exiting for accelerating unified vision language model}.
\newblock In \emph{{IEEE/CVF} Conference on Computer Vision and Pattern Recognition, {CVPR} 2023, Vancouver, BC, Canada, June 17-24, 2023}, pages 10781--10791. {IEEE}.

\bibitem[{Tang et~al.(2022)Tang, Han, Wang, Xu, Guo, Xu, and Tao}]{DBLP:conf/cvpr/Tang00XGXT22}
Yehui Tang, Kai Han, Yunhe Wang, Chang Xu, Jianyuan Guo, Chao Xu, and Dacheng Tao. 2022.
\newblock \href {https://doi.org/10.1109/CVPR52688.2022.01185} {Patch slimming for efficient vision transformers}.
\newblock In \emph{{IEEE/CVF} Conference on Computer Vision and Pattern Recognition, {CVPR} 2022, New Orleans, LA, USA, June 18-24, 2022}, pages 12155--12164. {IEEE}.

\bibitem[{Vaswani et~al.(2017)Vaswani, Shazeer, Parmar, Uszkoreit, Jones, Gomez, Kaiser, and Polosukhin}]{transformer}
Ashish Vaswani, Noam Shazeer, Niki Parmar, Jakob Uszkoreit, Llion Jones, Aidan~N Gomez, \L~ukasz Kaiser, and Illia Polosukhin. 2017.
\newblock \href {https://proceedings.neurips.cc/paper/2017/file/3f5ee243547dee91fbd053c1c4a845aa-Paper.pdf} {Attention is all you need}.
\newblock In \emph{Advances in Neural Information Processing Systems}, volume~30. Curran Associates, Inc.

\bibitem[{Wang et~al.(2021)Wang, Zhang, and Han}]{SpAtten}
Hanrui Wang, Zhekai Zhang, and Song Han. 2021.
\newblock \href {https://doi.org/10.1109/HPCA51647.2021.00018} {Spatten: Efficient sparse attention architecture with cascade token and head pruning}.
\newblock In \emph{{IEEE} International Symposium on High-Performance Computer Architecture, {HPCA} 2021, Seoul, South Korea, February 27 - March 3, 2021}, pages 97--110. {IEEE}.

\bibitem[{Wang et~al.(2020{\natexlab{a}})Wang, Hu, Zhang, Li, Wang, Zhang, Gao, and Liu}]{minivlm}
Jianfeng Wang, Xiaowei Hu, Pengchuan Zhang, Xiujun Li, Lijuan Wang, Lei Zhang, Jianfeng Gao, and Zicheng Liu. 2020{\natexlab{a}}.
\newblock \href {http://arxiv.org/abs/2012.06946} {Minivlm: {A} smaller and faster vision-language model}.
\newblock \emph{CoRR}, abs/2012.06946.

\bibitem[{Wang et~al.(2022{\natexlab{a}})Wang, Zheng, Chen, and Wang}]{redundancy_in_vit}
Peihao Wang, Wenqing Zheng, Tianlong Chen, and Zhangyang Wang. 2022{\natexlab{a}}.
\newblock \href {https://openreview.net/forum?id=O476oWmiNNp} {Anti-oversmoothing in deep vision transformers via the fourier domain analysis: From theory to practice}.
\newblock In \emph{The Tenth International Conference on Learning Representations, {ICLR} 2022, Virtual Event, April 25-29, 2022}. OpenReview.net.

\bibitem[{Wang et~al.(2022{\natexlab{b}})Wang, Yang, Men, Lin, Bai, Li, Ma, Zhou, Zhou, and Yang}]{ofa}
Peng Wang, An~Yang, Rui Men, Junyang Lin, Shuai Bai, Zhikang Li, Jianxin Ma, Chang Zhou, Jingren Zhou, and Hongxia Yang. 2022{\natexlab{b}}.
\newblock \href {https://proceedings.mlr.press/v162/wang22al.html} {{OFA:} unifying architectures, tasks, and modalities through a simple sequence-to-sequence learning framework}.
\newblock In \emph{International Conference on Machine Learning, {ICML} 2022, 17-23 July 2022, Baltimore, Maryland, {USA}}, volume 162 of \emph{Proceedings of Machine Learning Research}, pages 23318--23340. {PMLR}.

\bibitem[{Wang et~al.(2023{\natexlab{a}})Wang, Zhou, Zeng, and Zhang}]{efficientvlm}
Tiannan Wang, Wangchunshu Zhou, Yan Zeng, and Xinsong Zhang. 2023{\natexlab{a}}.
\newblock \href {https://doi.org/10.18653/V1/2023.FINDINGS-ACL.873} {Efficientvlm: Fast and accurate vision-language models via knowledge distillation and modal-adaptive pruning}.
\newblock In \emph{Findings of the Association for Computational Linguistics: {ACL} 2023, Toronto, Canada, July 9-14, 2023}, pages 13899--13913. Association for Computational Linguistics.

\bibitem[{Wang et~al.(2023{\natexlab{b}})Wang, Bao, Dong, Bjorck, Peng, Liu, Aggarwal, Mohammed, Singhal, Som, and Wei}]{beit3}
Wenhui Wang, Hangbo Bao, Li~Dong, Johan Bjorck, Zhiliang Peng, Qiang Liu, Kriti Aggarwal, Owais~Khan Mohammed, Saksham Singhal, Subhojit Som, and Furu Wei. 2023{\natexlab{b}}.
\newblock \href {https://doi.org/10.1109/CVPR52729.2023.01838} {Image as a foreign language: {BEIT} pretraining for vision and vision-language tasks}.
\newblock In \emph{{IEEE/CVF} Conference on Computer Vision and Pattern Recognition, {CVPR} 2023, Vancouver, BC, Canada, June 17-24, 2023}, pages 19175--19186. {IEEE}.

\bibitem[{Wang et~al.(2020{\natexlab{b}})Wang, Wei, Dong, Bao, Yang, and Zhou}]{minilmv1}
Wenhui Wang, Furu Wei, Li~Dong, Hangbo Bao, Nan Yang, and Ming Zhou. 2020{\natexlab{b}}.
\newblock \href {https://proceedings.neurips.cc/paper/2020/hash/3f5ee243547dee91fbd053c1c4a845aa-Abstract.html} {Minilm: Deep self-attention distillation for task-agnostic compression of pre-trained transformers}.
\newblock In \emph{Advances in Neural Information Processing Systems 33: Annual Conference on Neural Information Processing Systems 2020, NeurIPS 2020, December 6-12, 2020, virtual}.

\bibitem[{Wang et~al.(2022{\natexlab{c}})Wang, Wang, Zhu, Liu, Qin, and Wei}]{dide}
Zekun Wang, Wenhui Wang, Haichao Zhu, Ming Liu, Bing Qin, and Furu Wei. 2022{\natexlab{c}}.
\newblock \href {https://doi.org/10.18653/V1/2022.EMNLP-MAIN.608} {Distilled dual-encoder model for vision-language understanding}.
\newblock In \emph{Proceedings of the 2022 Conference on Empirical Methods in Natural Language Processing, {EMNLP} 2022, Abu Dhabi, United Arab Emirates, December 7-11, 2022}, pages 8901--8913. Association for Computational Linguistics.

\bibitem[{Wang et~al.(2020{\natexlab{c}})Wang, Wohlwend, and Lei}]{DBLP:conf/emnlp/WangWL20}
Ziheng Wang, Jeremy Wohlwend, and Tao Lei. 2020{\natexlab{c}}.
\newblock \href {https://doi.org/10.18653/v1/2020.emnlp-main.496} {Structured pruning of large language models}.
\newblock In \emph{Proceedings of the 2020 Conference on Empirical Methods in Natural Language Processing, {EMNLP} 2020, Online, November 16-20, 2020}, pages 6151--6162. Association for Computational Linguistics.

\bibitem[{Xia et~al.(2022)Xia, Zhong, and Chen}]{cofi}
Mengzhou Xia, Zexuan Zhong, and Danqi Chen. 2022.
\newblock \href {https://doi.org/10.18653/v1/2022.acl-long.107} {Structured pruning learns compact and accurate models}.
\newblock In \emph{Proceedings of the 60th Annual Meeting of the Association for Computational Linguistics (Volume 1: Long Papers), {ACL} 2022, Dublin, Ireland, May 22-27, 2022}, pages 1513--1528. Association for Computational Linguistics.

\bibitem[{Xie et~al.(2019)Xie, Lai, Doran, and Kadav}]{snli_ve}
Ning Xie, Farley Lai, Derek Doran, and Asim Kadav. 2019.
\newblock \href {http://arxiv.org/abs/1901.06706} {Visual entailment: {A} novel task for fine-grained image understanding}.
\newblock \emph{CoRR}, abs/1901.06706.

\bibitem[{Xin et~al.(2020)Xin, Tang, Lee, Yu, and Lin}]{DeeBert}
Ji~Xin, Raphael Tang, Jaejun Lee, Yaoliang Yu, and Jimmy Lin. 2020.
\newblock \href {https://doi.org/10.18653/v1/2020.acl-main.204} {Deebert: Dynamic early exiting for accelerating {BERT} inference}.
\newblock In \emph{Proceedings of the 58th Annual Meeting of the Association for Computational Linguistics, {ACL} 2020, Online, July 5-10, 2020}, pages 2246--2251. Association for Computational Linguistics.

\bibitem[{Xu et~al.(2020)Xu, Zhou, Ge, Wei, and Zhou}]{BERT-of-Theseus}
Canwen Xu, Wangchunshu Zhou, Tao Ge, Furu Wei, and Ming Zhou. 2020.
\newblock \href {https://doi.org/10.18653/v1/2020.emnlp-main.633} {Bert-of-theseus: Compressing {BERT} by progressive module replacing}.
\newblock In \emph{Proceedings of the 2020 Conference on Empirical Methods in Natural Language Processing, {EMNLP} 2020, Online, November 16-20, 2020}, pages 7859--7869. Association for Computational Linguistics.

\bibitem[{Xu et~al.(2021)Xu, Zhou, Fu, Zhou, and Li}]{xujingjing_survey}
Jingjing Xu, Wangchunshu Zhou, Zhiyi Fu, Hao Zhou, and Lei Li. 2021.
\newblock \href {http://arxiv.org/abs/2111.05193} {A survey on green deep learning}.
\newblock \emph{CoRR}, abs/2111.05193.

\bibitem[{Xu et~al.(2023)Xu, Wu, Rosenman, Lal, Che, and Duan}]{bridgetower}
Xiao Xu, Chenfei Wu, Shachar Rosenman, Vasudev Lal, Wanxiang Che, and Nan Duan. 2023.
\newblock \href {https://doi.org/10.1609/AAAI.V37I9.26263} {Bridgetower: Building bridges between encoders in vision-language representation learning}.
\newblock In \emph{Thirty-Seventh {AAAI} Conference on Artificial Intelligence, {AAAI} 2023, Thirty-Fifth Conference on Innovative Applications of Artificial Intelligence, {IAAI} 2023, Thirteenth Symposium on Educational Advances in Artificial Intelligence, {EAAI} 2023, Washington, DC, USA, February 7-14, 2023}, pages 10637--10647. {AAAI} Press.

\bibitem[{Xu et~al.(2022)Xu, Zhang, Zhang, Sheng, Li, Dong, Zhang, Xu, and Sun}]{evo_vit}
Yifan Xu, Zhijie Zhang, Mengdan Zhang, Kekai Sheng, Ke~Li, Weiming Dong, Liqing Zhang, Changsheng Xu, and Xing Sun. 2022.
\newblock \href {https://ojs.aaai.org/index.php/AAAI/article/view/20202} {Evo-vit: Slow-fast token evolution for dynamic vision transformer}.
\newblock In \emph{Thirty-Sixth {AAAI} Conference on Artificial Intelligence, {AAAI} 2022, Thirty-Fourth Conference on Innovative Applications of Artificial Intelligence, {IAAI} 2022, The Twelveth Symposium on Educational Advances in Artificial Intelligence, {EAAI} 2022 Virtual Event, February 22 - March 1, 2022}, pages 2964--2972. {AAAI} Press.

\bibitem[{Ye et~al.(2021)Ye, Lin, Huang, and Sun}]{TR-BERT}
Deming Ye, Yankai Lin, Yufei Huang, and Maosong Sun. 2021.
\newblock \href {https://doi.org/10.18653/v1/2021.naacl-main.463} {{TR-BERT:} dynamic token reduction for accelerating {BERT} inference}.
\newblock In \emph{Proceedings of the 2021 Conference of the North American Chapter of the Association for Computational Linguistics: Human Language Technologies, {NAACL-HLT} 2021, Online, June 6-11, 2021}, pages 5798--5809. Association for Computational Linguistics.

\bibitem[{Yin et~al.(2022)Yin, Vahdat, Alvarez, Mallya, Kautz, and Molchanov}]{A-vit}
Hongxu Yin, Arash Vahdat, Jose~M. Alvarez, Arun Mallya, Jan Kautz, and Pavlo Molchanov. 2022.
\newblock \href {https://doi.org/10.1109/CVPR52688.2022.01054} {A-vit: Adaptive tokens for efficient vision transformer}.
\newblock In \emph{{IEEE/CVF} Conference on Computer Vision and Pattern Recognition, {CVPR} 2022, New Orleans, LA, USA, June 18-24, 2022}, pages 10799--10808. {IEEE}.

\bibitem[{Yu et~al.(2022)Yu, Wang, Vasudevan, Yeung, Seyedhosseini, and Wu}]{coca}
Jiahui Yu, Zirui Wang, Vijay Vasudevan, Legg Yeung, Mojtaba Seyedhosseini, and Yonghui Wu. 2022.
\newblock \href {https://openreview.net/forum?id=Ee277P3AYC} {Coca: Contrastive captioners are image-text foundation models}.
\newblock \emph{Trans. Mach. Learn. Res.}, 2022.

\bibitem[{Zeng et~al.(2022)Zeng, Zhang, and Li}]{xvlm}
Yan Zeng, Xinsong Zhang, and Hang Li. 2022.
\newblock \href {https://proceedings.mlr.press/v162/zeng22c.html} {Multi-grained vision language pre-training: Aligning texts with visual concepts}.
\newblock In \emph{International Conference on Machine Learning, {ICML} 2022, 17-23 July 2022, Baltimore, Maryland, {USA}}, volume 162 of \emph{Proceedings of Machine Learning Research}, pages 25994--26009. {PMLR}.

\bibitem[{Zhou et~al.(2023)Zhou, Jiang, Cotterell, and Sachan}]{efficientprompt}
Wangchunshu Zhou, Yuchen~Eleanor Jiang, Ryan Cotterell, and Mrinmaya Sachan. 2023.
\newblock Efficient prompting via dynamic in-context learning.
\newblock \emph{CoRR}, abs/2305.11170.

\bibitem[{Zhou et~al.(2020)Zhou, Xu, Ge, McAuley, Xu, and Wei}]{pabee}
Wangchunshu Zhou, Canwen Xu, Tao Ge, Julian~J. McAuley, Ke~Xu, and Furu Wei. 2020.
\newblock \href {https://proceedings.neurips.cc/paper/2020/hash/d4dd111a4fd973394238aca5c05bebe3-Abstract.html} {{BERT} loses patience: Fast and robust inference with early exit}.
\newblock In \emph{Advances in Neural Information Processing Systems 33: Annual Conference on Neural Information Processing Systems 2020, NeurIPS 2020, December 6-12, 2020, virtual}.

\end{thebibliography}


\appendix

\section{Details of Similarity Calculation}
\label{sec:similarity}
To measure the redundancy in token representations and attention heads of VLMs, we calculate the average cosine similarity between token representations and attention maps at each layer following previous work~\citep{power_bert, redundancy_in_vit}.

\paragraph{Token Similarity}
Given the corresponding token representations $\mX \in \R^{N \times D}$, the averaged token representations similarity is computed by:
\begin{gather*}
    \mS_{\textit{T}} = \frac{2}{N(N-1)} \sum_{i=1}^{N} \sum_{j=i+1}^{N} \frac{\mX_{i} \cdot \mX_{j}}{\left\lVert \mX_{i} \right\rVert_2 \left\lVert \mX_{j} \right\rVert_2}
\end{gather*}

\paragraph{Head Similarity}
We use the similar metric to compute head similarity for attention maps.
Given the attention map $\mA \in \R^{H \times N \times N}$ with $H$ heads, the averaged cosine similarity between different heads is calculated as:
\begin{gather*}
    \mS_{\textit{A}} = \frac{2}{H(H-1)N} \sum_{i=1}^{H} \sum_{j=i+1}^{H} \sum_{k=1}^{N} \frac{\mA_{i}^{k} \cdot \mA_{j}^{k}}{{\big\lVert \mA_{i}^{k} \big\rVert}_2 {\big\lVert \mA_{j}^{k} \big\rVert}_2}
\end{gather*}
where $\mA_i^k$ denotes the $k$-th token's attention distribution in the $i$-th head.

\begin{figure}[t]
    \begin{center}
    \includegraphics[clip, width=\linewidth]{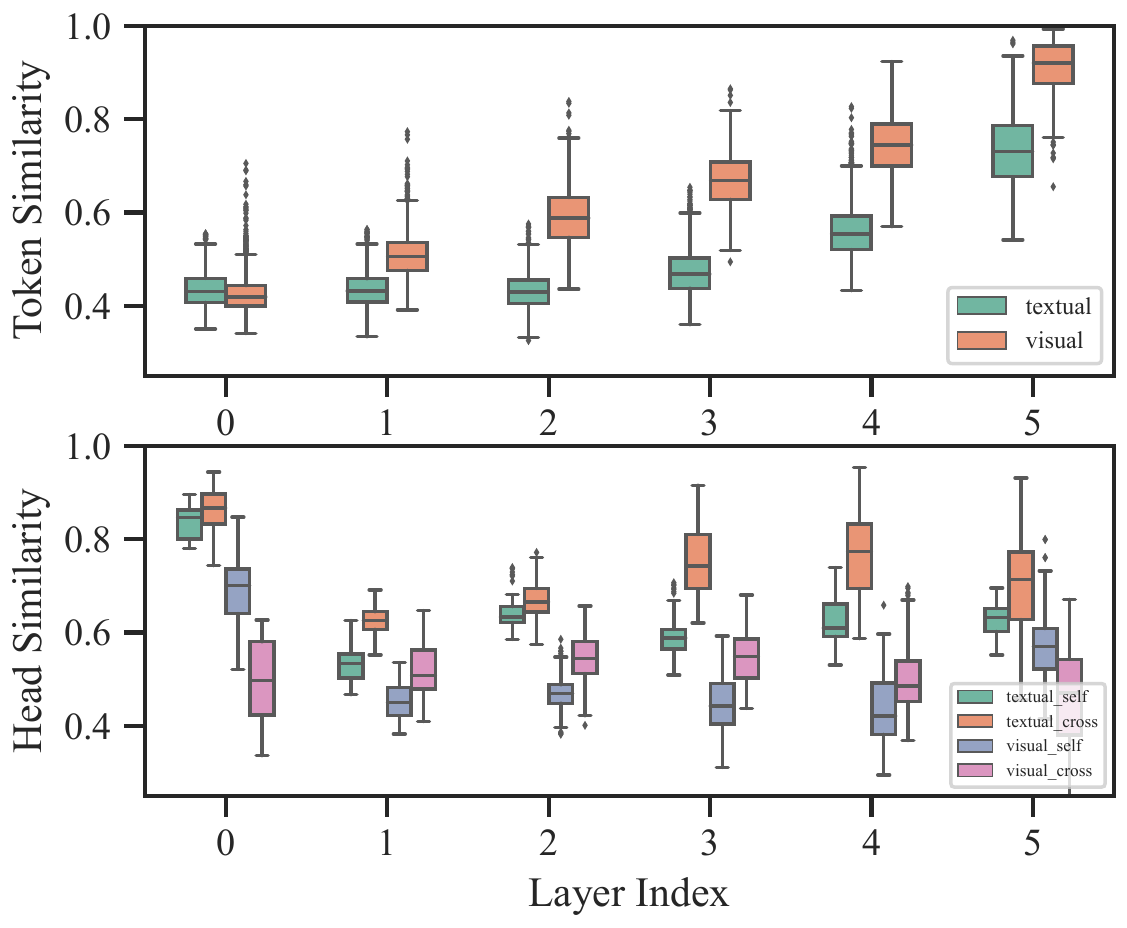}
    \caption{ 
        The similarity visualizations of the cross-modal encoder in METER fine-tuned on NLVR2.
    }
    \label{fig:nlvr2_cross_sim}
    \end{center}
\end{figure}

\begin{figure}[t]
    \begin{center}
    \includegraphics[clip, width=\linewidth]{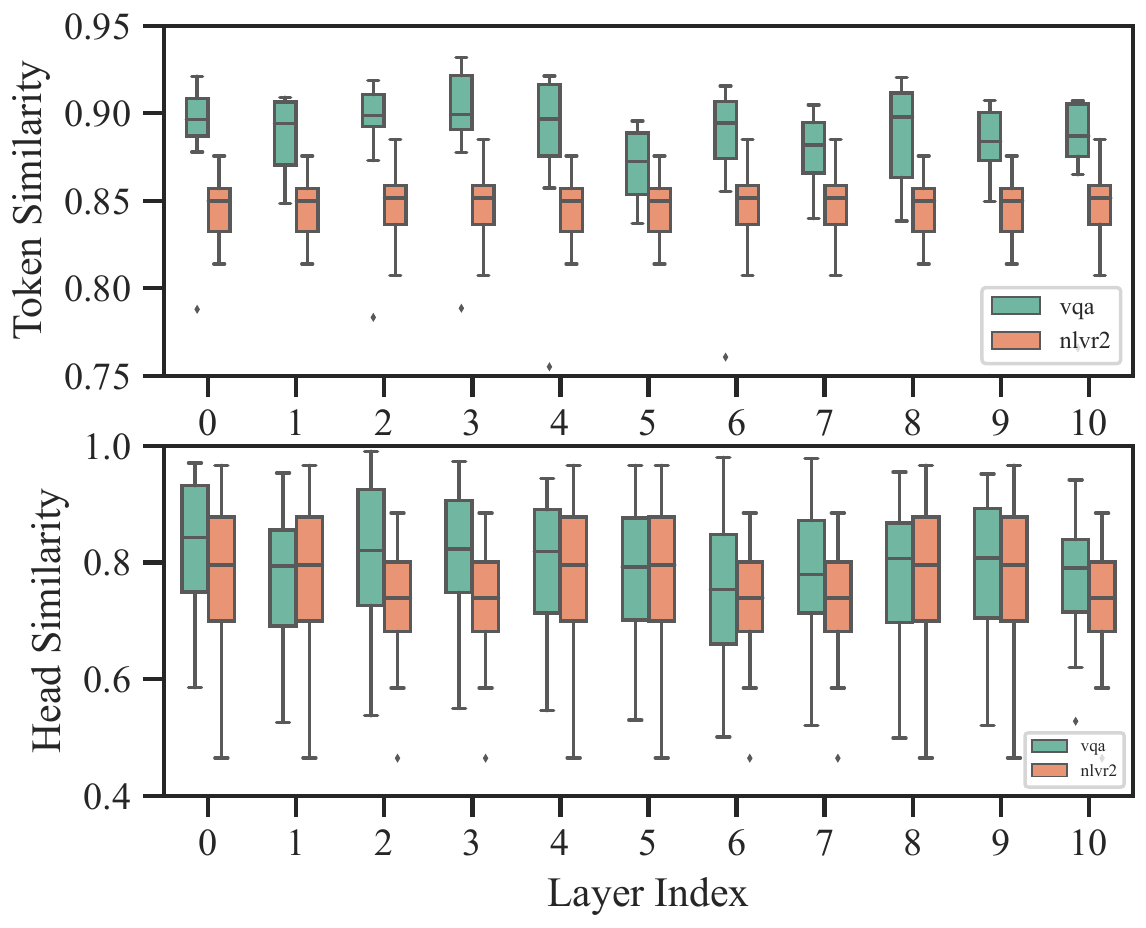}
    \caption{ 
        The similarity visualizations of the textual encoder in METER fine-tuned on VQA and NLVR2.
    }
    \label{fig:text_sim}
    \end{center}
\end{figure}

\begin{figure}[t]
    \begin{center}
    \includegraphics[clip, width=\linewidth]{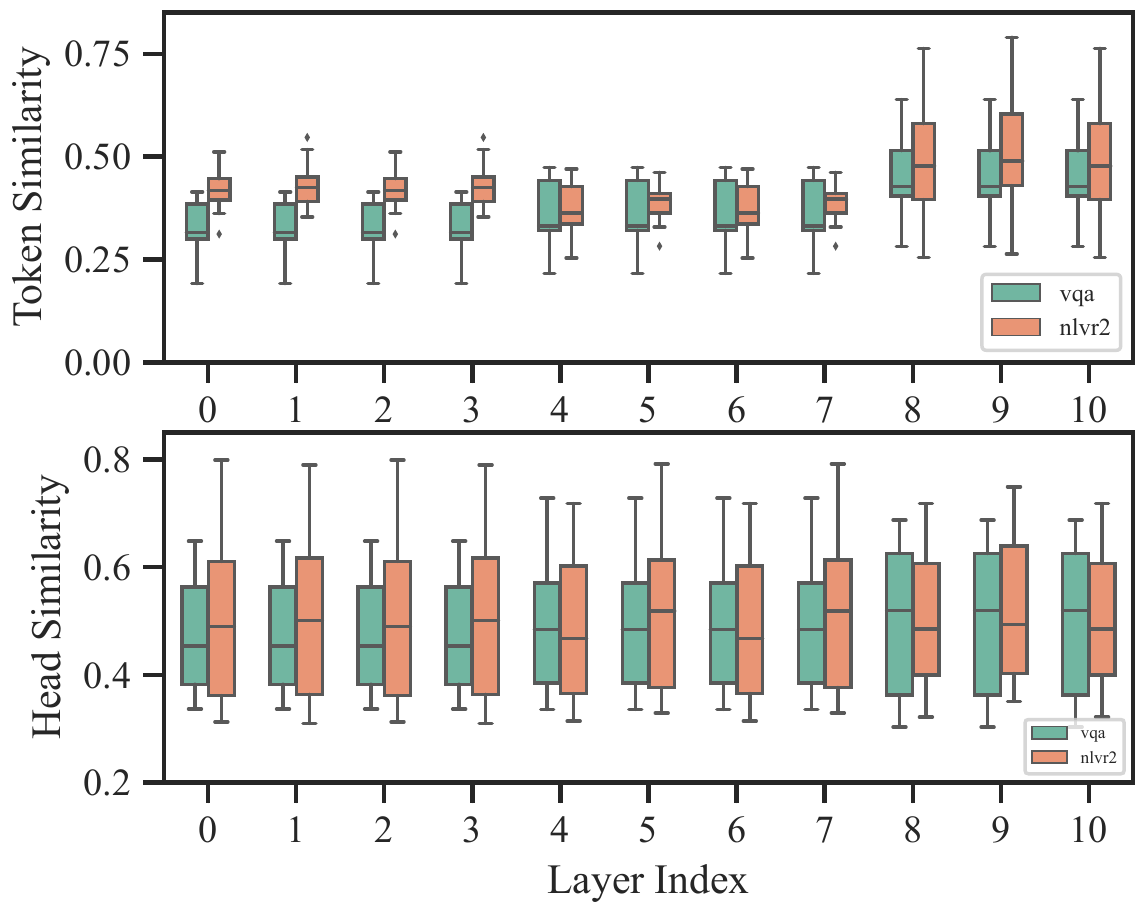}
    \caption{ 
        The similarity visualizations of the visual encoder in METER fine-tuned on VQA and NLVR2.
    }
    \label{fig:vit_sim}
    \end{center}
\end{figure}

\paragraph{More Visualization}
We also present the visualizations of different modules in VLMs on NLVR2 and VQA tasks in Figures~\ref{fig:nlvr2_cross_sim},~\ref{fig:text_sim}, and~\ref{fig:vit_sim}.
Similar to Figure~\ref{fig:vqa_cross_sim}, significant redundancy can be observed in both token representations and attention heads within the VLM modules on various tasks.

\section{Details of Downstream Tasks}
\label{sec:datasets}
\paragraph{Natural Language for Visual Reasoning} (NLVR2~\cite{NLVR2}) is a visual reasoning task that aims to determine whether a textual statement describes a pair of images.
For METER-based models, we construct two pairs of image-text, each consisting of the image and a textual statement.
For models based on BLIP, we directly feed the two images and the text to the encoder.

\paragraph{Visual Question Answering}(VQA v2~\citep{vqav2}) requires the model to answer questions based on the input image.
For METER-based models, we formulate the problem as a classification task with 3,129 answer candidates.
For BLIP-based models, we consider it as an answer generation task and use the decoder to rank the candidate answers during inference.

\paragraph{Visual Entailment}(SNLI-VE~\citep{snli_ve}) is a three-way classification dataset, aiming to predict the relationship between an image and a text hypothesis: \textit{entailment}, \textit{natural}, and \textit{contradiction}.

\paragraph{Image-Text Retrieval} (ITR)
We evaluate image-to-text retrieval (TR) and
text-to-image retrieval (IR) on Flickr30K~\citep{Flickr} with the standard split~\citep{Karpathy_split}.

\paragraph{Image Captioning}
The image is given to the encoder and the decoder will generate the corresponding caption with a text prompt "a picture of" following~\citet{blip}.
In this work, we optimize only the cross-entropy loss during fine-tuning.
Our experiments are conducted on COCO~\citep{coco_dataset}, and the evaluation is performed on both the COCO test set and the NoCaps~\citep{NoCaps} validation set (zero-shot transfer).

\section{Implementation Details}
\begin{table}[!t]
\centering
\small
\resizebox{\linewidth}{!}{
    \begin{tabular}{l | c c c c}
    \toprule
    Hyperparameters & NLVR2 & VQAv2 & SNLI-VE & Flickr30K \\
    \midrule
    Epochs & 10 & 10 & 5 & 10 \\
    Batch Size & 256 & 512 & 64 & 512 \\
    Initial Learning Rate & 1e-5 & 5e-6 & 2e-6 & 5e-6 \\
    Learning Rate Decay & \multicolumn{4}{c}{Linear Scheduler} \\
    Dropout & \multicolumn{4}{c}{0.1} \\
    Weight Decay & \multicolumn{4}{c}{0.01} \\
    Warmup Ratio & \multicolumn{4}{c}{0.1} \\
    AdamW $\beta$ & \multicolumn{4}{c}{(0.9, 0.999)} \\
    Data Augmentation & \multicolumn{4}{c}{RandomAugment} \\
    Image Resolution & \multicolumn{4}{c}{$288^{2}$} \\
    \bottomrule
    \end{tabular}
    }
    \caption{
    Hyperparameters for fine-tuning \our{}-METER on various downstream VL tasks.
    }
\label{tab:meter_params}
\end{table}
\begin{table}[!t]
\centering
\small
\resizebox{\linewidth}{!}{
    \begin{tabular}{l | c c c}
    \toprule
    Hyperparameters & NLVR2 & VQAv2 & Captioning \\
    \midrule
    Epochs & 15 & 10 & 5 \\
    Batch Size & \multicolumn{3}{c}{256} \\
    Initial Learning Rate & 3e-5 & 2e-5 & 1e-5 \\
    Learning Rate Decay & \multicolumn{3}{c}{Cosine Scheduler} \\
    Weight Decay & \multicolumn{3}{c}{0.05} \\
    AdamW $\beta$ & \multicolumn{3}{c}{(0.9, 0.999)} \\
    Data Augmentation & \multicolumn{3}{c}{RandomAugment} \\
    Image Resolution & $384^{2}$ & $480^{2}$ & $384^{2}$ \\
    \bottomrule
    \end{tabular}
    }
    \caption{
    Hyperparameters for fine-tuning \our{}-BLIP on various downstream VL tasks.
    }
\label{tab:blip_params}
\end{table}
\subsection{Hyperparameter Settings}
\label{sec:details_imp}
The MLP network in our token and head trimmers consists of two linear layers with GeLU activation~\citep{gelu}.
To reduce the computations, we set $D^{\prime} = D / 12$.
Fine-tuning hyperparameters on METER are given in Table~\ref{tab:meter_params}, mainly following the defaults in \citet{meter}.
Fine-tuning hyperparameters on BLIP are given in Table~\ref{tab:blip_params}, mainly following the defaults in \citet{blip}.
We perform token adaptive pruning in the visual encoder/cross-modal encoder and head adaptive pruning in the cross-modal encoder.
For efficiency evaluation, we use \textit{torchprofile} to measure FLOPs.
As for the latency, we evaluate on an Intel Xeon E5-466 2640 v4 CPU.

\subsection{Details of Re-implemented Baselines}
\label{sec:details_baselines}
For FTKD, we initiate the student model following~\citet{pkd-bert} to directly use the first $k$ layers of the original model ($k \in \{4,6\}$ for the visual encoder, $k \in \{2,3\}$ for the cross-modal encoder).
In our experiments, we find that this initialization strategy is considerably better than the other methods.
Then, we fine-tune the student model by logit/hidden representation/attention distillation objectives the same as \citet{tinybert}.
For MuE, we fine-tune the METER according to~\citet{MuE}, and perform grid search from $0.85$ to $0.99$, an interval of 0.01, for the similarity thresholds of the visual and cross-modal encoder.
For TRIPS, we follow the original setting in \citet{TRIPS} to fine-tune the METER backbone.
We exhaustively search for optimal settings and hyperparameters for the re-implemented baselines.

\subsection{Details of Baselines for Trimming Ablation}
\label{sec:details_ablations}
Here we provide details of baselines in the trimming ablation.
\paragraph{Token Trimming}
For the \textit{local} baseline, we remove the cross-modal awareness score when calculating the token importance.
The \textit{random} baseline randomly prunes tokens during both training and inference.
Following previous work~\citep{power_bert,evit,TRIPS}, the \textit{Attn} baseline adopts the token attention value as the importance score and uses top-\textit{k} operation to select retained tokens, discarding the remaining ones.
For a fair comparison, we ensure that all baselines incur the same computational overhead as our method. 
In addition, we conduct an exhaustive search to determine the optimal hyperparameters for each baseline. 
This meticulous approach ensures the comparability of our method with other methods.

\paragraph{Head Trimming}
For a given retention ratio $p\%$,
the random baseline randomly retains $p\%$ of heads in each attention module.
Gradient-based head pruning~\citep{DBLP:conf/nips/MichelLN19} first computes loss on pseudo-labels and then prunes attention heads with the importance score obtained by Taylor expansion. 
With given input $x$, importance score of head $h$ is defined as:
\begin{gather*}
    \mI_{\textit{h}} = E_x \left| A_h^T \frac{\partial \mathcal{L}(x)}{\partial A_h}\right|
\end{gather*}
Where $\mathcal{L}$ is the loss function, and $A_h$ is the context layer of head $h$.
For the gradient-based baseline, we introduce two variants:
(1) \textit{Grad Local}, which retains the top-$p\%$ heads in each attention module,
(2) \textit{Grad All}, which maintains the top-$p\%$ heads of the entire model.
We apply these methods on the METER cross-modal encoder.

\begin{figure*}[!ht]
    \begin{center}
    \includegraphics[clip, width=\linewidth]{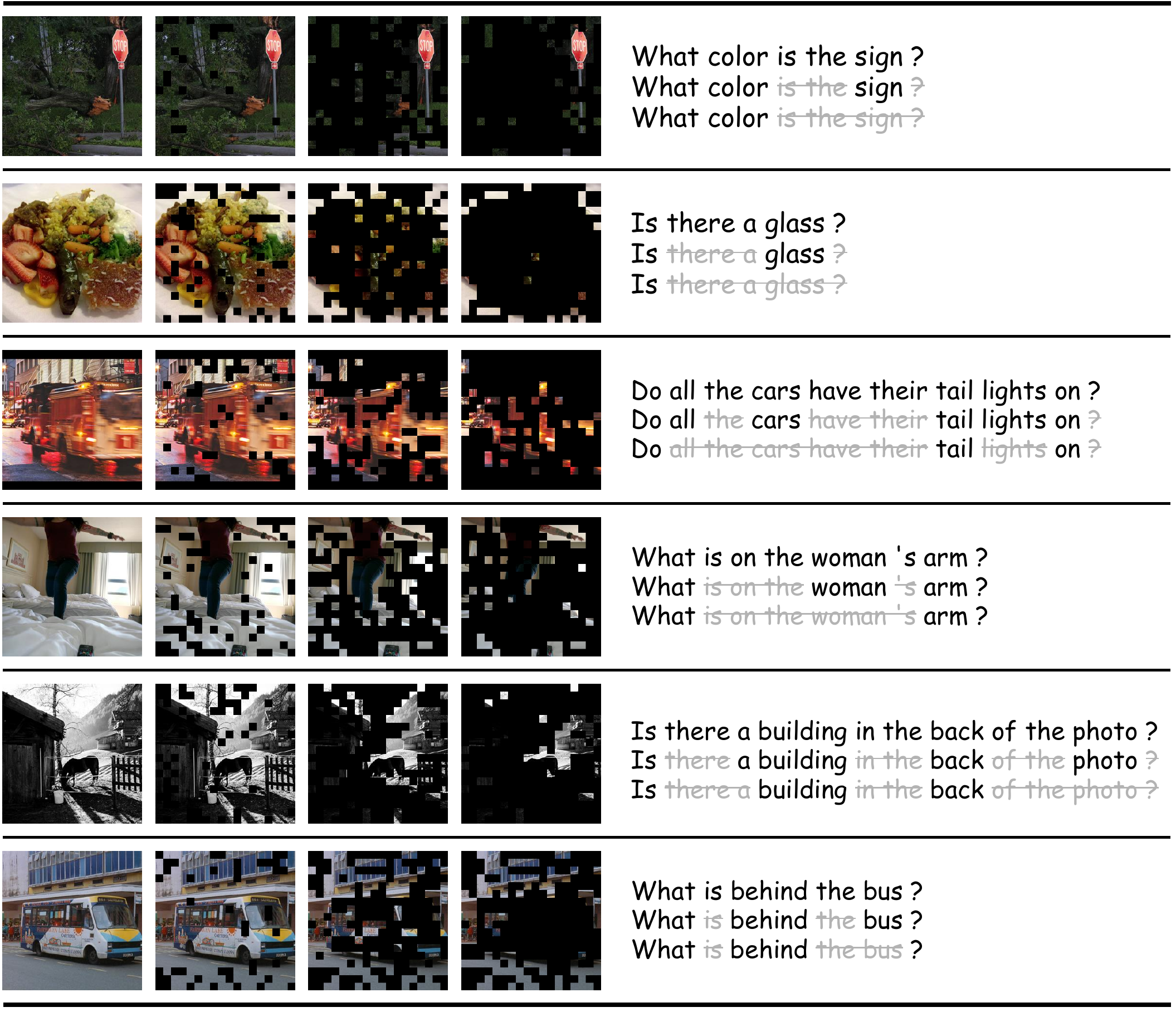}
    \caption{
    More visualization results by~\our{}.
    }
    \label{fig:more_visualization}
    \end{center}
\end{figure*}

\section{More Visualization Examples of Token Trimming}
\label{sec:more_visualization}
To demonstrate the ability to understand cross-modal interactions of our approach, we show more visualization results of our XModal-aware token trimmer in Figure \ref{fig:more_visualization}.
We can see that the final retained image patches are highly relevant to the textual questions.
The question words (e.g., \textit{what}) are critical in VQA because they are highly correlated with the category (numbers, yes/no or others) of correct answers.
Therefore, we observe that function words (e.g., \textit{of,the}) are gradually removed while critical tokens such as question words are retained.

\end{document}